\documentclass[preprint,5p,sort&compress,times]{elsarticle}

\usepackage{lineno,hyperref}
\modulolinenumbers[5]

\journal{Journal of Web Semantics}

\usepackage{times}%
\usepackage{dcolumn}
\usepackage{graphics}
\usepackage{graphicx}
\usepackage{url}
\usepackage{longtable}
\usepackage{supertabular}
\usepackage{pdflscape}
\usepackage{pifont}

\usepackage{rotating}
\usepackage{array}
\usepackage{tabularx}
\usepackage{lscape}
\usepackage{multirow}
\usepackage{multicol}
\usepackage{pdflscape}
\usepackage{float}
\usepackage{amsthm}

\usepackage{mathtools}
\usepackage{footnote}
\usepackage{verbatim}
\usepackage{supertabular}
\usepackage{hyperref}

\usepackage{url}
\usepackage{mathtools}
\usepackage{scrextend}
\usepackage{xspace}
\usepackage[textwidth=50pt]{todonotes}
\usepackage{listingsutf8}
\usepackage{color}
\usepackage{booktabs}
\usepackage[nolist]{acronym}
\usepackage{tikz}
\usepackage{lipsum}
\usepackage[numbers]{natbib}
\usepackage{tabulary}
\usepackage{svg}
\usepackage{threeparttable, tablefootnote}
\usepackage{fourier}

\definecolor{gray}{rgb}{0.4,0.4,0.4}
\definecolor{darkblue}{rgb}{0.0,0.0,0.6}
\definecolor{cyan}{rgb}{0.0,0.6,0.6}
\definecolor{cffffff}{RGB}{255,255,255}

\lstdefinelanguage{XML}
{
  morestring=[b]",
  morestring=[s]{>}{<},
  morecomment=[s]{<?}{?>},
  stringstyle=\color{black},
  identifierstyle=\color{darkblue},
  keywordstyle=\color{cyan},
  morekeywords={xmlns,version,type}
}

\definecolor{codegreen}{rgb}{0,0.6,0}
\definecolor{codegray}{rgb}{0.5,0.5,0.5}
\definecolor{codeblue}{rgb}{0,0,255}
\definecolor{backcolour}{rgb}{0.95,0.95,0.92}

\lstdefinestyle{rdf}{numberblanklines=true, morekeywords={},
backgroundcolor=\color{backcolour},   
    commentstyle=\color{codegreen},
    keywordstyle=\color{magenta},
    numberstyle=\tiny\color{codegray},
    stringstyle=\color{codeblue},
    basicstyle=\footnotesize,
    breakatwhitespace=false,         
    breaklines=true,                 
    captionpos=b,                    
    keepspaces=true,                 
    numbers=left,                    
    numbersep=5pt,                  
    showspaces=false,                
    showstringspaces=false,
    showtabs=false,                  
    tabsize=2
}

\lstdefinestyle{sparql}{numberblanklines=true, morekeywords={SERVICE,SELECT,DISTINCT,SAMPLE,FROM,WHERE,FILTER,ORDER,GROUP,BY,IN,AS,LIMIT}}
 
\newcolumntype{d}[1]{D{.}{.}{#1}}
\newcommand{\tablefont}[1]{\fontsize{3mm}{3.2mm}\selectfont}

\newcommand{\executeiffilenewer}[3]{%
 \ifnum\pdfstrcmp{\pdffilemoddate{#1}}%
 {\pdffilemoddate{#2}}>0%
 {\immediate\write18{#3}}\fi%
}

\begin{document}
\begin{acronym}[UML]
	\acro{AOS}{Agricultural Ontology Services}
	\acro{AGRIS}{Agricultural Science and Technology}
	\acro{API}{Application Programming Interface}
	\acro{BPSO}{Binary Particle-Swarm Optimization}
	\acro{BPMLOD}{Best Practices for Multilingual Linked Open Data}
	\acro{CBD}{Concise Bounded Description}
	\acro{COG}{Content Oriented Guidelines}
	\acro{CSV}{Comma-Separated Values}
	\acro{CBMT}{Corpus-Based Machine Translation}
	\acro{CLIR}{Cross-Language Information Retrieval}
	\acro{DPSO}{Deterministic Particle-Swarm Optimization}
	\acro{DALY}{Disability Adjusted Life Year}

	\acro{ER}{Entity Resolution}
	\acro{EM}{Expectation Maximization}
	\acro{EBMT}{Example-Based Machine Translation}
	\acro{EBNF}{Extended Backus--Naur Form}
	\acro{EL}{Entity Linking}
	\acro{FAO}{Food and Agriculture Organization of the United Nations}
	\acro{GIS}{Geographic Information Systems}
	\acro{GHO}{Global Health Observatory}
	\acro{HDI}{Human Development Index}
	\acro{ICT}{Information and communication technologies}
    \acro{KB}{Knowledge Base}
    \acro{KBSE}{Knowledge Base Semantic Embedding}
	\acro{LR}  {Language Resource}
	\acro{LD}  {Linked Data}
	\acro{LLOD}  {Linguistic Linked Open Data}
	\acro{LIMES}{LInk discovery framework for MEtric Spaces}
	\acro{LS}  {Link Specifications}
	\acro{LDIF}{Linked Data Integration Framework}
	\acro{LGD} {LinkedGeoData}
	\acro{LOD} {Linked Open Data}
	\acro{MSE}{Mean Squared Error}
	\acro{MWE}{Multiword Expressions}
	\acro{MT}{Machine Translation}
	\acro{ML}{Machine Learning}
	\acro{NIF}{Natural Language Processing Interchange Format}
	\acro{NIF4OGGD}{NLP Interchange Format for Open German Governmental Data}
	\acro{NLP}{Natural Language Processing}
	\acro{NER}{Named Entity Recognition}
	\acro{NMT}{Neural Machine Translation}
	\acro{NN}{Neural Network}
	\acro{NLG}{Natural Language Generation}
	\acro{NED}{Named Entity Disambiguation}
	\acro{NERD}{Named Entity Recognition and Disambiguation}
	\acro{NL}{Natural Language}
	\acro{OSM}{OpenStreetMap}
	\acro{OWL}{Web Ontology Language}
	\acro{OOV}{out-of-vocabulary}
	\acro{PFM}{Pseudo-F-Measures}
	\acro{PSO}{Particle-Swarm Optimization}
	\acro{QA}{Question Answering}
	\acro{RDF}{Resource Description Framework}
	\acro{RBMT}{Ru\-le-Ba\-sed Ma\-chi\-ne Trans\-la\-tion}
	\acro{SKOS}{Simple Knowledge Organization System}
	\acro{SPARQL}{SPARQL Protocol and RDF Query Language}
	\acro{SRL}{Statistical Relational Learning}
	\acro{SWT}{Semantic Web Technologies}
	\acro{SW}{Semantic Web}
	\acro{SMT}{Statistical Machine Translation}
	\acro{SWMT}{Semantic Web Machine Translation}

    \acro{TBMT} {Transfer-Based Machine Translation}
	\acro{UML}{Unified Modeling Language}
	\acro{USL}{Ukrainian Sign Language}
	\acro{WHO}{World Health Organization}
	\acro{WKT}{Well-Known Text}
	\acro{W3C}{World Wide Web Consortium}
	\acro{WSD}{Word Sense Disambiguation}
    \acro{XML}{Extensible Markup Language}
	\acro{YPLL}{Years of Potential Life Lost}
\end{acronym}  
\begin{frontmatter}                           

%
\title{Machine Translation using Semantic Web Technologies: A Survey}




\author[leipzig,pade]{Diego Moussallem \corref{cor}}
\ead{moussallem@informatik.uni-leipzig.de}

\author[leipzig]{Matthias Wauer}
\ead{wauer@informatik.uni-leipzig.de}

\author[pade]{Axel-Cyrille Ngonga Ngomo}
\ead{axel.ngonga@upb.de}

\address[leipzig]{University of Leipzig\\ 
AKSW Research Group\\
Department of Computer Science\\
Augustusplatz 10, 04109 Leipzig, Germany}

\address[pade]{University of Paderborn \\
Data Science Group\\
Pohlweg 51, D-33098 Paderborn, Germany}

\cortext[cor]{Principal corresponding author}

\begin{abstract}

A large number of machine translation approaches have recently been developed to facilitate the fluid migration of content across languages. However, the literature suggests that many obstacles must still be dealt with to achieve better automatic translations. One of these obstacles is lexical and syntactic ambiguity. A promising way of overcoming this problem is using Semantic Web technologies. This article presents the results of a systematic review of machine translation approaches that rely on Semantic Web technologies for translating texts. Overall, our survey suggests that while Semantic Web technologies can enhance the quality of machine translation outputs for various problems, the combination of both is still in its infancy.
\end{abstract}


\end{frontmatter}


\section{Introduction}
\label{sec:intro}
Alongside increasing globalization comes a greater need for readers to understand texts in languages foreign to them. For example, approximately 48\% of the pages on the Web are not available in English\footnote{\url{https://www.internetworldstats.com/stats7.htm}}.
The technological progress of recent decades has made both the distribution and access to content in different languages ever simpler. Translation aims to support users who need to access content in a language in which they are not fluent ~\cite{slocum1985survey, Koehn2010}.

However, translation is a difficult task due to the complexity of natural languages and their structure~\cite{Jurafsky2000}. In addition, manual translation does not scale to the magnitude of the Web. One remedy for this problem is \ac{MT}. The main goal of \ac{MT} is to enable people to assess content in languages other than the languages in which they are fluent~\cite{Bar-Hillel1960}. From a formal point of view, this means that the goal of \ac{MT} is to transfer the semantics of text from an input language to an output language~\cite{hutchins1992introduction}. At the time of writing, large information portals such as Google\footnote{http://translate.google.com.br/about/} or Bing\footnote{http://www.bing.com/translator/help/} already offer \ac{MT} services that are widely used. 

Although \ac{MT} systems are now popular on the Web, they still generate a large number of incorrect translations. Recently, Popovi{\'c}~\citep{popovic2012class} has classified five types of errors that still remain in \ac{MT} systems. According to  research, the two main faults that are responsible for 40\% and 30\% of problems respectively, are reordering errors and lexical and syntactic ambiguity. Thus, addressing these barriers is a key challenge for modern translation systems. 

A large number of \ac{MT} approaches have been developed over the years that could potentially serve as a remedy. For instance, translators began by using methodologies based on linguistics which led to the family of \ac{RBMT}. However, \ac{RBMT} systems have a critical drawback
in their reliance on manually crafted rules, thus making the development of new translation modules for different languages even more difficult~\cite{costa2012study,thurmair2004comparing}.

\ac{SMT} and \ac{EBMT} were developed to deal with the scalability issue in \ac{RBMT}~\cite{brown1990statistical}, a necessary characteristic of \ac{MT} systems that must deal with data at Web scale. Presently, these approaches have begun to address the drawbacks of rule-ba\-sed approaches. However, some problems that had already been solved for linguistics based methods reappeared. The majority of these problems are connected to the issue of ambiguity, including syntactic and semantic variations~\cite{Koehn2010}. 

Subsequently, \ac{RBMT} and \ac{SMT} have been combined in order to resolve the drawbacks of these two fa\-mi\-lies of approa\-ches. This combination of methods is called hybrid \ac{MT}. Although hybrid approaches have been achieving good results, they still suffer from some \ac{RBMT} problems~\cite{costa2015latest,costa2015much,Thumair}, for example, the big effort of adding new rules for handling a given syntax divergence. Nowadays, a novel \ac{SMT} paradigm has ari\-sen called \ac{NMT} which relies on \ac{NN} algorithms. \ac{NMT} has been achieving impressive results and is now the state-of-the-art in \ac{MT} approaches. However, \ac{NMT} is still a statistical approach sharing some semantic drawbacks from other well-defined \ac{SMT} approaches\citep{koehn2017six}.  

One possible solution to address the remaining issues of \ac{MT} lies in the use of \ac{SWT}, which have emerged over recent decades as a paradigm to make the semantics of content explicit so that it can be used by machines~\cite{Berners-Lee2001}. It is believed that explicit semantic knowledge made available through these technologies can empower \ac{MT} systems to supply translations with significantly better quality while remaining scalable~\cite{heuss2013lessons}. In particular, the disambiguated knowledge about real-world entities, their properties and their relationships made available on the \ac{LD} Web can potentially be used to infer the right meaning of ambiguous sentences or words and also to support the reordering task. 

The obvious opportunity of using \ac{SWT} for \ac{MT} has already been studied by a number of approaches. This systematic survey gives an overview of existing systems making use of this combination and presents the difference in translation quality that they produce, especially w.r.t. the issue of ambiguity. Based on this overview, we distill the challenges and opportunities in the use of \ac{SWT} in \ac{MT} for translating texts.

This paper is structured as follows: In~\autoref{sec:method}, we describe the methodology used to conduct this systematic survey.~\autoref{sec:mtapproaches} discusses different \ac{MT} approaches and their particular challenges.~\autoref{sec:mtsw} shows how \ac{SWT} have been used in \ac{MT} approaches and presents suggestions on how to handle the challenges.~\autoref{sec:conc} concludes with ideas for future work.

\section{Research Method}
\label{sec:method}
The research methodology behind this survey follows the formal systematic literature review process. In particular, this study is based on the guidelines proposed in~\cite{Dyba2007,Kitchenham2004,Moher:2009}. As detailed below, we also took into account other surveys from relevant journals as well as surveys about related topics such as \ac{WSD} and \ac{SW}.

\subsection{Research Questions}
\label{sec:researchquestions}

The goal of this survey is to provide \ac{SW} researchers with existing methodologies that use \ac{SWT} applied to \ac{MT} systems for translating natural-language sentences. To achieve this goal, we aimed to answer the following general research question: \emph{How can \ac{SWT} enhance \ac{MT} quality?}
This question was then divided into four sub-questions as follows:

\begin{itemize}

\item[RQ1.] What are state-of-the-art approaches in \ac{MT} which use \ac{SWT}?

\item[RQ2.] Which \ac{SWT} are applied in \ac{MT}?

\item[RQ3.] Does ontological knowledge influence the quality of an automatic translation?

\item[RQ4.] What kinds of \ac{SW} driven tools are available for \ac{MT}?

\end{itemize}

\textbf{RQ1} intends to collect available research works which retrieve knowledge from \ac{SW} resources for translating texts. 
\textbf{RQ2} aims to provide an explicit comparison among \ac{SWT} used in each respective \ac{MT} approach. \textbf{RQ3} attempts to resolve whether inclusion of a certain concept represented or inferred by an ontology supports and improves the translation process of a given \ac{MT} system. \textbf{RQ4} asks for a description of all available \ac{SW} tools that have been used and may be used in future work for supporting \ac{MT} systems.    

\subsection{Search Strategy}
\begin{figure*}[htb]
\centering
\includegraphics[scale=0.50]{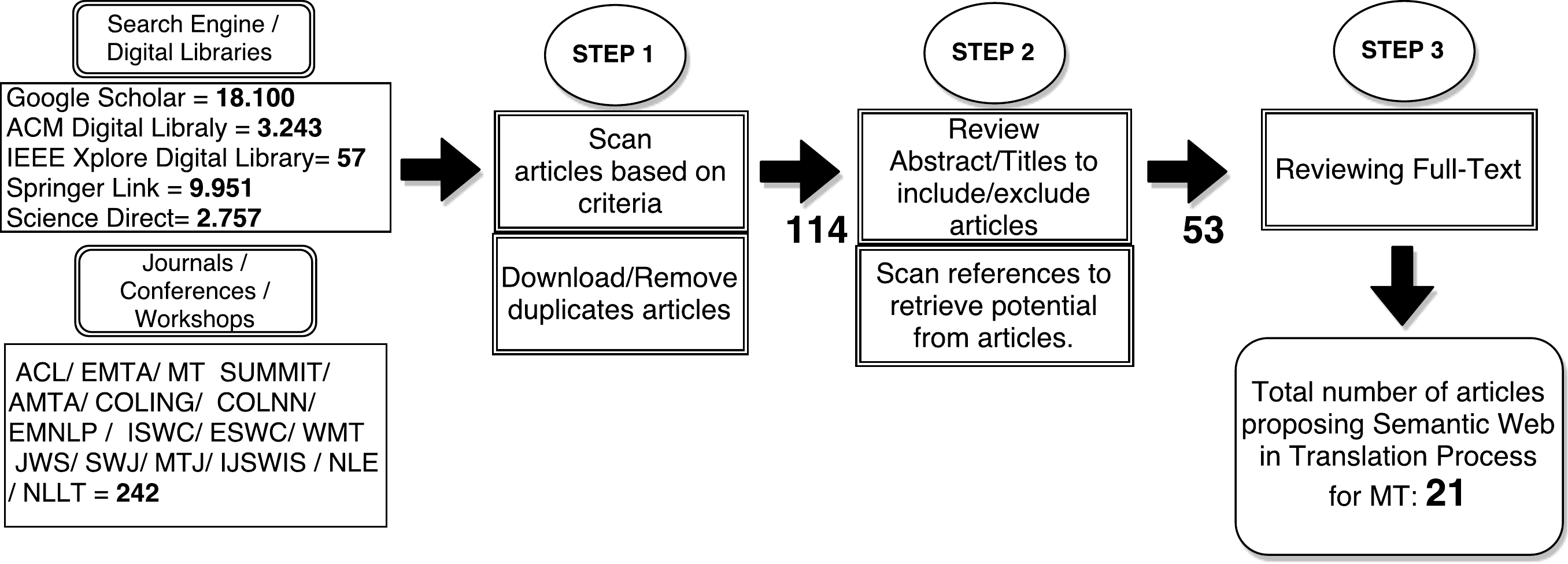}
\caption{Steps followed to retrieve the systematic review results.}
\label{fig:generalS}
\end{figure*}

An overview of our search methodology and the number of articles collected at each step is shown in~\autoref{fig:generalS} and described in detail below. To start the search, it was essential to determine search criteria that fit the purposes of our survey. Based on best practices~\cite{Dyba2007,Kitchenham2004,Moher:2009}, we defined the following selection criteria to classify the retrieved studies.

\subsubsection{Inclusion criteria} 

The papers considered in our study were publications in English between 2001 and 2017. They had to satisfy at least one of the following criteria:

\begin{itemize}

\item a focus on distinguishing between ambiguous words in \ac{MT} that use \ac{SWT}.


\item proposed and/or implemented an approach for \ac{MT} using \ac{SWT}.

\item contain a combination of ontological knowledge with \ac{MT} for handling structural divergence issues.

\end{itemize}

\subsubsection{Exclusion criteria} 
None of the criteria below was to hold for the papers considered in this survey:

\begin{itemize}

\item not peer-reviewed or published.

\item assessment methodologies published as a poster abstract.

\item no use of \ac{SWT} in \ac{MT} for translating natural language sentences.

\item not proposing an \ac{MT} approach or framework which retrieves information using \ac{SWT}.

\end{itemize}

\subsubsection{Search queries}

To address the research questions and criteria, we determined a set of keyword queries that allowed us to detect relevant studies for our survey. We used the following keywords: \textit{machine translation}, \textit{semantic web}, \textit{ontology}, \textit{linked data}, \textit{disambiguation}, \textit{methodology}, and \textit{multilingual}. The choice of these keywords corresponds to the main keywords used by \ac{SW} and \ac{MT} works in their titles. Also, the keywords \textit{multilingual} and \textit{methodology} were identified to return papers matching the inclusion criteria. Therefore, they were combined in two search queries in order to retrieve the relevant research works.\footnote{ The search engines automatically take into account the inflections and synonyms of the keywords.}

\begin{enumerate}

\item machine translation AND (ontology OR linked data OR semantic web OR disambiguation)

\item (methodology OR multilingual OR disambiguation) AND (linked data OR ontology OR semantic web OR machine translation)

\end{enumerate}

Thereafter, we used the following search engines, digital libraries, journals, conferences, and workshops to find relevant publications.

Search engines and digital libraries:
\begin{itemize}
\item Google Scholar\footnote{\url{http://scholar.google.com/}}
\item ACM Digital Library\footnote{\url{http://dl.acm.org/}}
\item IEEE Xplore Digital Library\footnote{\url{http://ieeexplore.ieee.org/}}
\item SpringerLink\footnote{\url{http://link.springer.com/}}
\item ScienceDirect\footnote{\url{http://www.sciencedirect.com/}}
\item MT-Archive\footnote{\url{http://www.mt-archive.info/}}
\end{itemize}
Journals:
\begin{itemize}
\item Semantic Web Journal
(SWJ)\footnote{\url{http://www.semantic-web-journal.net/}}
\item Journal of Web Semantics
(JWS)\footnote{\url{http://www.websemanticsjournal.org/}}
\item Machine Translation Journal
(MT)\footnote{\url{http://link.springer.com/journal/10590}}
\item Natural Language and Linguistic Theory\footnote{\url{http://link.springer.com/journal/11049}}
\item Natural Language Engineering\footnote{\url{http://journals.cambridge.org/action/displayJournal?jid=NLE}}
\item International Journal on Semantic Web and Information Systems 
(IJSWIS)\footnote{\url{http://www.ijswis.org/}}
\end{itemize}
Conferences and associated workshops:
\begin{itemize}
\item Association for Computational Linguistics (ACL)
~\cite{acl2017,acl2016,acl2015,acl2014,acl2013,acl2012,acl2011,acl2010,acl2009,acl2008,acl2007,acl2006,acl2005,acl2004,acl2003,acl2002}
\item North American Association for Computational Linguistics 
(NAACL)~\cite{naacl2016,naacl2015, naacl2013,naacl2012,naacl2010,naacl2009,naacl2007,naacl2006,naacl2005,naacl2004,naacl2003}
\item Empirical Methods in Natural Language Processing \-(EMNLP)
~\cite{emnlp2017,emnlp2016,emnlp2015,emnlp2014,emnlp2013,emnlp2012,emnlp2011,emnlp2010,emnlp2009,emnlp2008,emnlp2007,emnlp2006,emnlp2005,emnlp2004}
\item International Conference on Computational Linguistics (COLING)
~\cite{coling2016,coling2014,coling2012,coling2010,coling2008,coling2006,coling2004,coling2002}
\item International Association for Machine Translation (IAMT, AMTA, EAMT, MT Summit)\footnote{\url{http://www.mt-archive.info/srch/conferences-1.htm}}
\item World Wide Web Conference (WWW)
~\cite{www2017,www2016,www2011,www2012,www2013,www2014,www2015, www2010,www2009,www2008,www2007,www2006,www2005,www2004,www2003,www2002,www2001}
\item International Semantic Web Conference (ISWC)
~\cite{iswc2002,iswc2003,iswc2004,iswc2005,iswc2006,iswc2007,iswc2008,iswc2009,iswc2010,iswc2011,iswc2012,iswc2013,iswc2014,iswc2015,iswc2015, iswc2016,iswc2017}
\item Extended Semantic Web Conference (ESWC)
~\cite{eswc2017,eswc2016,eswc2011,eswc2012,eswc2013,eswc2014,eswc2015,eswc2010,eswc2009,eswc2008,eswc2007,eswc2006,eswc2005,eswc2004}

\end{itemize}

\subsection{Search Steps}

The initial search based on the queries described above returned 34,350  publications. The subsequent processing of this set of publications was divided into three steps.

\subsubsection{Step 1} This step ensures that the publications considered in the study abide by the selection criteria. Hence, we excluded publications that are not in English or do not contain any reference to \ac{SW} and \ac{MT}. Moreover, we excluded all publications that were not peer-reviewed, as well as works that were published as posters or abstracts. We manually scanned the articles based on the criteria presented above.

Thereafter, the bibliographic metadata of remaining publications was analyzed using the bibliography management platform Mendeley\footnote{\url{https://www.mendeley.com/groups/7405201/semantic-web-machine-translation/papers/}} and 21 duplicates were removed. 

\subsubsection{Step 2} For this step, we reviewed the abstracts of the 114 articles returned by Step 1 based on the four research questions described in \autoref{sec:researchquestions}. 64 of the 114 articles were excluded because they did not \ac{SWT}. However, all 114 articles were documented as a reference source for potential future articles. To retrieve articles from references we adopted the following strategy: 

\begin{enumerate}

\item We searched for the article title in Google Scholar and retrieved the ``Cited By'' articles.

\item We then read the abstract of each potential article found by searching the references of all 50 articles and by these means, found 3 more articles. 
\end{enumerate}

\subsubsection{Step 3} With a total of 53 articles deemed potentially relevant for the survey, we read all articles completely to evaluate their suitability for our study. As a final result, we selected 21 articles published between 2001 and 2017, which are listed in~\autoref{selectedpapers}. 10 of these 21 articles matched the criteria, which propose and/or implement \ac{MT} using \ac{SWT}. 6 of 21 matched the criteria for disambiguation processes in \ac{MT} using \ac{SWT}.  5 of 21 matched the criteria that focused on combining ontological knowledge with \ac{MT} to handle syntactic divergences. Most of the remaining relevant articles are related to the translation of ontology labels. Although ontology label translation is an important task relevant to some of the articles, we do not consider them in this survey because ontology labels are single words or compounds rather than sentences.

\begin{table*}[htb]
\centering
\caption{List of the selected papers.} 
\label{selectedpapers}
\scriptsize
\begin{tabular}{@{} l|l @{}}
\toprule
\textbf{Citation} & \textbf{Title}\\
\hline
C. Vertan~\cite{vertan2004language} & Language Resources for the Semantic Web - perspectives for Machine Translation\\
\hline
W. Hahn and C. Vertan~\cite{hahn2005challenges} & Challenges for the Multilingual Semantic Web\\
\hline
N. Elita and A. Birladeanu~\cite{elita2005first} & A First Step in Integrating an EBMT into the Semantic Web\\
\hline
C. Shi and H. Wang~\cite{shi2005research} & Research on Ontology-driven Chinese-English Machine Translation\\
\hline
N. Elita and M. Gavrila~\cite{elita2006enhancing} & Enhancing translation memories with semantic knowledge\\
\hline
E. Seo et al.~\cite{seo2009syntactic} & Syntactic and Semantic English-Korean Machine Translation using Ontology\\
\hline
P. Knoth et al.~\cite{knoth2010facilitating} & Facilitating Cross-Language Retrieval and Machine Translation by Multilingual Domain Ontologies\\
\hline
L. Lesmo et al.\cite{lesmo2011ontology} & An Ontology based Architecture for Translation\\
\hline
A. M. Almasoud and H. S. Al-Khalifa~\cite{almasoud2011proposed} & A Proposed Semantic Machine Translation System for Translating Arabic Text to Arabic Sign Language\\
\hline
B. Harriehausen-M\"uhlbauer and T. Heuss~\cite{harriehausen2012semantic,heuss2013lessons} & Semantic Web based Machine Translation\\
\hline
K. Nebhi et al.~\cite{nebhi2013nerits} & NERITS - A Machine Translation Mashup System using Wikimeta and Linked Open Data\\
\hline
J. P. McCrae and P. Cimiano~\cite{mccrae2013mining} & Mining Translations from the Web of Open Linked Data\\
\hline
D. Moussallem and R. Choren~\cite{moussallem2015using} & Using Ontology-based Context in the Portuguese-English Translation of Homographs in Textual Dialogues \\
\hline
O. Lozynska and M. Davydov~\cite{lozynska2015information} & Information technology for Ukrainian Sign Language translation based on Ontologies\\
\hline
K. Simov et al.~\cite{simovtowards} & Towards Semantic-based Hybrid Machine Translation between Bulgarian and English \\
\hline
T.S. Santosh Kumar.~\cite{kumar2016} & Word Sense Disambiguation Using Semantic Web for Tamil to English Statistical Machine Translation\\
\hline
N. Abdulaziz et al.~\cite{neama2016towards} & Towards an Arabic-English Machine-Translation Based on Semantic Web\\
\hline
J. Du et al.~\cite{jinhua2016} & Using BabelNet to Improve OOV Coverage in SMT\\
\hline
A. Srivastava et al.~\cite{srivastava2016} & How to Configure Statistical Machine Translation with Linked Open Data Resources\\
\hline
C. Shi et al.~\cite{shi2016knowledge} & Knowledge-based Semantic Embedding for Machine Translation. \\
\hline
A. Srivastava et al.~\cite{srivastava2017} & Improving Machine Translation through Linked Data. \\
\bottomrule
\end{tabular}
\end{table*}

\section{Classification of MT Approaches}
\label{sec:mtapproaches}
In this section, we give an overview of generic dimensions across which \ac{MT} systems can be classified. This overview allows for a better understanding of the approaches retrieved as described above. A detailed description of the approaches is given in ~\autoref{sec:architectures}. Sequentially, we introduce the remaining dimensions. Afterwards, we present the open \ac{MT} challenges pertaining to all \ac{MT} approaches. Finally, we briefly introduce common \ac{MT} evaluation metrics in order to provide a background to how \ac{MT} systems are evaluated automatically.

\subsection{Dimensions}
We classify \ac{MT} systems across three dimensions.

\begin{enumerate}
\item \emph{Architecture}: From an architectural perspective, 
it is assumed that all \ac{MT} paradigms could be subsumed under one architecture to model existing \ac{MT} systems~\cite{Koehn2010}. However, the architecture may be composed of more than one approach and these approaches depend, for their operation, on the amount of available knowledge. For example, some approaches rely purely on statistics (\ac{SMT}) while others use complex linguistic models (\ac{RBMT}) to compute a translation.

\item \emph{Problem space addressed}: Previous works (e.g., \cite{Pushpak2015}) suggest that particular \ac{MT} approaches are best suited to address particular types of problems. For instance, an approach for translating old Egyptian texts should rely on deep linguistics rules due to the lack of bilingual corpora. In contrast, translating large volumes of text is best carried out using statistics models because resolving many errors from hand-crafted rules requires a big human effort. Furthermore, the usage of statistics will depend on the language and availability of bilingual corpora for training. 

\item \emph{Performance}: A central challenge of \ac{MT} is to create real-time \ac{MT} solutions that achieve a high-quality translation of variable-nature texts, while being low in complexity to build. For example, while \ac{SMT} performs well on long texts (paragraphs containing 100 words), it often fails on short sentences ( e.g., social network comments and subtitles) - mainly when a given \ac{SMT} is built using large corpora and needs to translate text from different domains~\cite{Koehn2010}.
\end{enumerate}

\subsection{Architectures}
\label{sec:architectures}

\begin{figure*}[htb]
\centering
\includegraphics[scale=0.40]{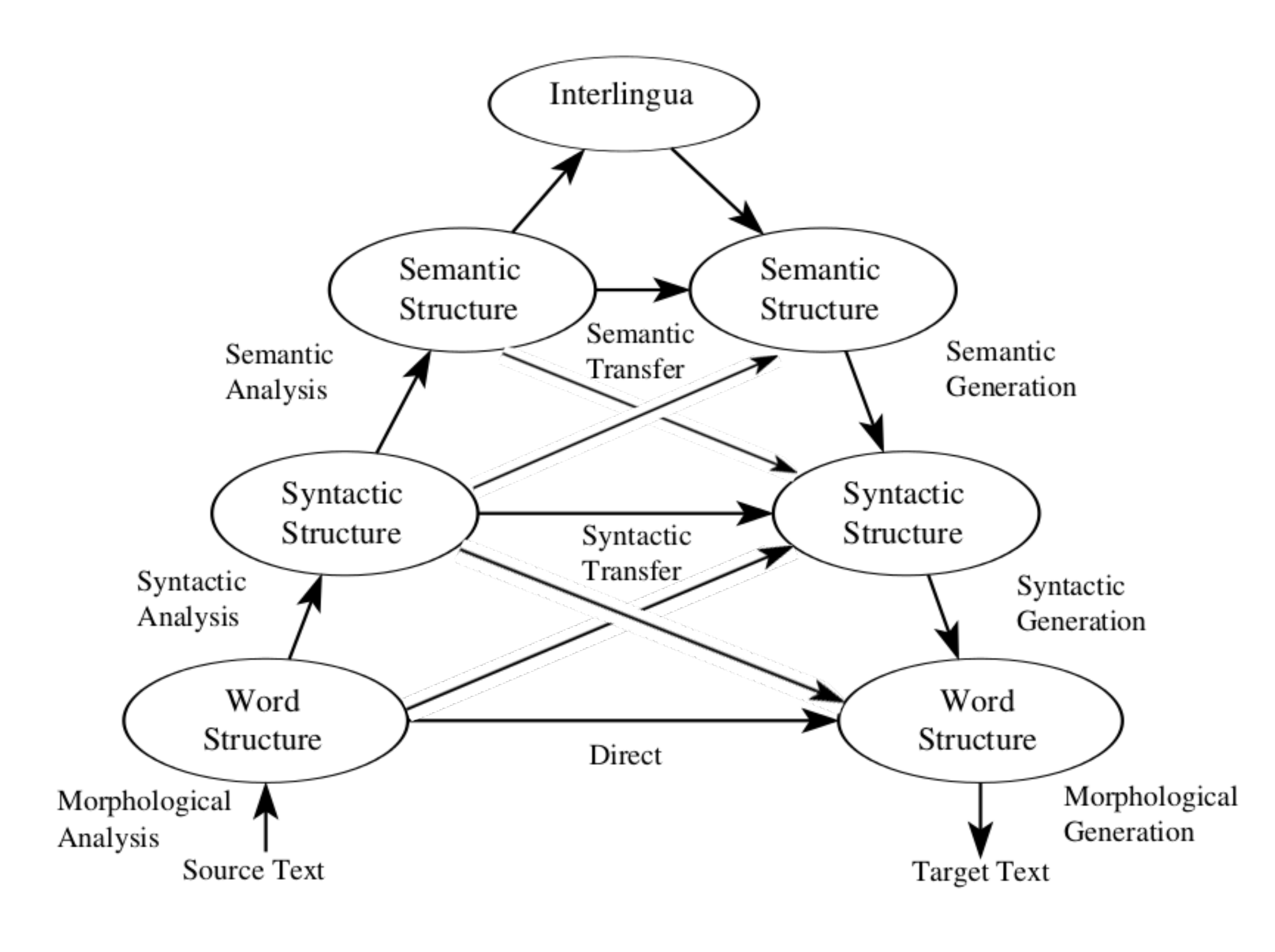}
\caption{Generic \ac{MT} Architecture - Vauquois triangle}
\label{fig:mtarch}
\end{figure*}

Currently, the architectures of \ac{MT} systems are subdivided into \ac{RBMT}, \ac{SMT} and \ac{EBMT}. In addition, hybrid systems which combine \ac{RBMT} with \ac{SMT} have emerged over recent years. \autoref{fig:mtarch} gives an overview of a generic \ac{MT} architecture that may represent the workflow of all approaches. The left side of the triangle comprises generically of source text analysis, while the right side corresponds to the generation of target texts. Both sides of this generic architecture have four main steps. (1) A morphological step, which handles the morphology of words. (2) A syntactic step, which deals with the structure of sentences. (3) A semantics' step, which considers the meaning of words and sentences. Finally, (4) an interlingual phase, which may be seen as a generic representation of source and target text either in \ac{RBMT} (internal model)  or \ac{SMT} approaches (statistical model). 

\subsubsection{Rule-Based Machine Translation}

At a basic level, \ac{RBMT} approaches carry out a translation in the following manner: first, they parse the input text. Then they create an intermediary linguistic representation of the scanned text. They finish by generating text in the target language based on morphological, syntactic, and semantic mappings between the two languages. \ac{RBMT} approaches can be divided into three categories: direct, transfer-based, and interlingua-based. In order to perform well, all of these approaches require extensive lexicons and large sets of rules designed by professional linguists. 

\paragraph{Direct approaches} This family emerged with the first \ac{MT} created by IBM~\cite{Brown:1993:MSM:972470.972474}. Another translator which implemented this idea was Systran~\cite{li2013survey}. Commonly, direct approaches are word-based or dictionary-based, translating the words one by one. Hence, they translate the texts without considering the meaning variations in the words, which leads to a significant error rate. Thus, the direct translation is seen as the first step in modern \ac{MT} systems and since it can easily be combined with other technologies, it is commonly used by hybrid \ac{MT} systems. 

\paragraph{Transfer-based approaches} \ac{TBMT} was created due to the obvious necessity of preserving meaning whilst translating. \ac{TBMT} has three steps: analysis, transfer, and generation. In the first step, \ac{TBMT} systems analyze the sentence structure of the source language and generate an internal representation based on linguistic patterns of the target language. Subsequently, \ac{TBMT} systems translate only the text for both respective languages. To this end, they use three dictionaries during the translation process: two monolingual dictionaries (source language and target language) and a bilingual dictionary containing a mapping between the source and the target languages~\cite{cheragui2012theoretical}.

\paragraph{Interligua-based approaches} The interlingual appro\-ach relies on similar insights to those underlying \ac{TBMT}. The main difference between them is the internal model that they rely upon. In contrast to \ac{TBMT} approaches (where the target language plays a key role), Interlingua-based approaches rely on extracting a language-independent representation of the input text. Hence, this can be easily adapted to translation into multiple target languages. State-of-the-art interlingual approaches rely most commonly on morphological, syntactic, and semantic rules based on a grammatical framework~\cite{ranta2011grammatical}. These are more accurate than comparable previously mentioned approaches (i.e., direct and \ac{TBMT}). In addition, a direct beneficial consequence of the interlingual approach is the time-efficient portability of its framework to other languages. This approach is widely used by linguists for morphologically rich languages because it allows them to choose the better language representation of a translation model ba\-sed on the source text. On the other hand, to achieve a  high-quality translation with Interlingua, the operator must carry out the translation within a specific domain.

\subsubsection{Statistical Machine Translation} 

\ac{SMT} relies upon the intuition that \ac{RBMT} shortcomings can be addressed using experience. Commonly,~\ac{SMT} approa\-ches acquire this experience from previous translations. These previous translations are collected from bilingual corpora, either parallel or comparable. They are crucial for the performance of \ac{SMT} approaches. A bilingual parallel corpus is comprised of a source text and its respective translation. In contrast, a comparable corpus does not contain an exact translation of a source text, but a similar corpus in another language which may share the same topic, size or given time. \ac{SMT} approaches heavily depend on \ac{ML} techniques including supervised, unsupervised, and semi-super\-vised algorithms to compute a statistical translation model from the input corpus for use in subsequent translations. Most commonly, the goal of \ac{SMT} approaches is simply to detect the translation that achieves the maximum probability or score for a given translation pair. \ac{SMT} requires well-suited parallel or comparable bilingual corpora to achieve high accuracy. Comparable corpora can be used to get more terminological knowledge into a certain \ac{SMT} system, but the \ac{SMT} still needs an underlying language model built on parallel data. Hence, the translation quality directly depends on the language model trained on a given target side of the parallel corpus, which is a subset of the respective bilingual corpora.

Every \ac{SMT} system shares at least three common steps. First, an alignment between the parallel or comparable corpora is necessary. Both source and target corpora are aligned in order to provide a mapping between each respective translation. When parallel corpora do not exist, there are ways to handle comparable corpora using algorithms that take alignment probabilities into account. Although it may affect the quality of a given \ac{SMT}, this is a common task when there are insufficient reference translations. Second, a translation model needs to be created and trained over the alignment previously done, thus being responsible for translating the content from source to target language. At training phase, an underlying model is created which contains statistical frequencies of words and may contain syntactic information (from a hierarchy model or manually inserted) about both languages. Once the model has been created, the source text can be processed by using a function akin to a probability or score distribution contained in the model. Afterward, the step called decoding is essentially a search function that accesses the translation model by getting all the translation candidates and providing a ranked list of possible translations. The language model which is created from a reference translation usually made by humans guarantees the fluency of a certain target language~\cite{Koehn2010}. The following five sub-approaches exist in \ac{SMT}.

\textbf{Word-based} - This approach was the first statistical one created by IBM which contains at least five IBM models. The \emph{Word-based} approach basically translates one word at a time based on its frequency computed by the translation model over the entire training data. By ignoring multiple meanings of a given word (e.g., polysemous or homonymous words), the approach can generate wrong translations. Considering this weakness, Yamada and Knight\cite{yamada2001syntax} then introduced the \textbf{Phrase-based} approach, which was subsequently improved by Koehn~\cite{koehn2003statistical}. It translates based on phrases corresponding to more than one word. Thus, this approach is able to consider the surrounding context of each word when dealing with ambiguous words. The \emph{Phrase-based} approach is considered a huge advancement with respect to \ac{MT} approaches. However, problems such as reordering and ambiguity still exist especially when translating multi-domain contents, which basically do not have a vocabulary overlap. This lack of vocabulary is known as the \ac{OOV} problem, \ac{OOV} words are unknown words which appear while testing a given \ac{MT} approach but do not occur when training the model.  

Regarding the remaining errors, one remedy found was to include linguistic rules in the translation model. Consequently, the \textbf{Syntax-based} (also known as Factored models) and \textbf{Tree-based} approaches appeared. They combine linguistics rules with statistical models, but the linguistic rules used by both approaches are different from those in \ac{RBMT}. The difference lies in the limitations imposed by applying some complex linguistic rules in a polynomial efficient time. Thus, some crucial linguistic rules are not considered. The factored models use part-of-speech tagging, which is responsible for decreasing structural problems while \emph{Tree-based} models attempt to resolve the reordering problem by using syntactic trees between languages. This combination solves basic and important problems caused by the structure of languages in \emph{Phrase-based MT}. In spite of \emph{Tree-based} approaches showing notable improvements, they skip a large number of necessary parsing rules, thus continuing to produce material errors~\cite{Koehn2010}.

Recently, a new statistical approach called \textbf{Neural Network-based} (also known as \ac{NMT})\footnote{see new chapter at Koehn's book~\url{http://statmt.org/mtma16/uploads/mtma16-neural.pdf}} has gained significant popularity~\cite{kalchbrenner2013recurrent,bahdanau2014neural}. Although \ac{NN} techniques are not essentially novel, they have been widely exploited since Google published their research~\cite{sutskever2014sequence} on beneficial improvements applying \ac{NN} on se\-quen\-ce-to-sequence models.\footnote{\url{http://www.statmt.org/survey/Topic/NeuralNetworkModels}} The structure of a given \ac{NMT} system is much simpler than the well-known Phrase-based \ac{SMT} in terms of components. In \ac{NMT} systems, there are no separate models (e.g., language model, translation mo\-del), rather, they usually consist of one sequence model  responsible for encoding and decoding the words. This se\-quen\-ce model learns a context vector created from the sources and target sentences for translating the texts. The context vector, i.e., embeddings, is an important part as it represents the continuous space representation of words and allows the sequence model to predict similar words. Despite some researchers claiming that \ac{NMT} may practically solve many of the weaknesses of \ac{SMT} systems, it is still  recent research and, as a statistical approach, still suffers from some well-known statistical problems~\cite{koehn2017six} that are described in \autoref{sec:problems}.


\subsubsection{Example-based Machine Translation} 

Similar to statistical approaches, \ac{EBMT} use bilingual corpora, but they store the data by example sentences. The training phase of \ac{EBMT} approaches imitates a basic memory technique (analogy) of human translators and is akin to filling the database of a translation memory system~\cite{king2001method}. Here, \ac{EBMT} approaches split the source text of a training corpus into sentences and align each of these sentences with the corresponding target sentences from the same training dataset. When an \ac{EBMT} is used, the known source sentences are fetched first. If any of these sentences occur in the input, the machine replaces it by the corresponding translation. If during the translation an unknown word occurs in a sentence (i.e., a word which was not seen in the training data), the word is not translated. This is clearly the main limitation of \ac{EBMT} approaches, which is addressed by hybrid MT systems. If several translations are available, the translation process chooses the target sentence based on a ranking algorithm. As \ac{EBMT} systems are related to memory techniques, they require large physical memory to store the translated sentences. Although the \ac{MT} community has moved forward with this approach, it is quite interesting and easily combined with the others.
 
\subsection{Problem space addressed by MT}
\label{sec:problems}

According to~\cite{Pushpak2015}, some \ac{MT} approaches may perform better than others for translating contents depending on the problem, language and how much bilingual content is available. Although \ac{RBMT} approaches are still currently used due to the difficulty of finding bilingual or comparable corpora containing syntactic information for some morphologically rich languages, the \ac{MT} community has practically moved on from \ac{RBMT} approaches because of the complicated extension of their linguistic rules. It has occurred due to the exponential number of research works on \ac{SMT}, based on languages which are widely spoken around the world and consequently have more bilingual data, such as English, French and Chinese. Thus, the creation of hand-crafted rules from these languages is no longer needed, as the \ac{ML} algorithms seemed to resolve the problems. 

Additionally, Koehn~\cite{Koehn2010} states that acquiring comparable corpora is made simple by using crawling techniques on the Web. However, the process of alignment required by \ac{SMT} approaches is substantially harder, and even more so without an explicit well-defined Web format. For instance, there is a task created by the well-known \ac{MT} workshop with the goal of aligning sentences automatically from Web pages in order to create comparable corpora\footnote{\url{http://www.statmt.org/wmt16/bilingual-task.html}}. The results and the training phase of this task confirm the difficulty of the process, for example, the reasonable difference between possible and training pairs where the training pairs are 1,624 and possible pairs are 225,043. Possible pairs are pairs which exist from a human perspective and may be found by an algorithm. In contrast, training pairs are accurate pairs which are given for training the candidate algorithms. 
Although the results of this task are promising, the evaluated systems require improvements for creating a real \ac{MT} training data.  Also, most comparable corpora are limited in terms of domains (e.g politics, technology and financial).
   
\subsection{MT performance and effort}
\label{sec:efforts}

All \ac{MT} approaches have advantages and disadvantages w.r.t. quality and efforts. In~\autoref{Comparison}, we compare the approaches presented in \autoref{sec:architectures}, considering the dimensions stated previously.

\begin{table*}[htb]
\centering
\begin{threeparttable}
\small
\caption{MT Approaches Comparison}
\label{Comparison}
\begin{tabular}{@{} l|l @{}}
\toprule
\multicolumn {2}{c}{\textbf{Rule-based}} \\
\hline
\textbf{Pros} & \textbf{Cons} \\
 \hline
 Deep linguistic knowledge (\textbf{quality})~\tnote{a} & Requires linguistic rules and dictionaries (\textbf{effort})~\tnote{b} \\
 \hline
 Easy to perform error analysis (\textbf{effort})~\tnote{b} &  Human language inconsistency (\textbf{quality})~\tnote{a} \\
 \hline
 &  Expensive to extend and improve (\textbf{effort})~\tnote{b} \\
 \hline
\multicolumn {2}{c}{\textbf{Statistical}} \\ 
\hline
\textbf{Pros} &\textbf{Cons} \\
\hline
No complex processing of linguistic rules (\textbf{effort})~\tnote{b} & Require parallel text~\tnote{c}~~ (\textbf{quality}~\tnote{a}~~,~\textbf{effort}~\tnote{b}~~~) \\
 \hline
 Less human resources cost, no linguists required (\textbf{effort})~\tnote{b} & No linguistic rules causes syntactic/semantic errors (\textbf{quality})~\tnote{a} \\
 \hline
 Applicable to any pair of languages (\textbf{effort})~\tnote{b} & Difficult to perform error analysis (\textbf{effort})~\tnote{b}~~, especially for \ac{NMT} \\
 \hline
 Models trained with human translations (\textbf{quality})~\tnote{a} & Preprocessing noisy training data (\textbf{effort})~\tnote{b} \\
\bottomrule
\end{tabular}
\begin{tablenotes}\footnotesize
\item [\textbf{a}] Translation quality: In Pros, it means to improve the quality of outputs while it means to reduce the quality in Cons.
\item [\textbf{b}] Human efforts: In Pros, it means to decrease the human efforts while in Cons, it means to increases the human efforts.
\item [\textbf{c}] some \ac{NMT} approaches have started to use only monolingual data for an unsupervised  training~\cite{lample2017unsupervised}.
\end{tablenotes}
\end{threeparttable}
\end{table*}

The main weakness of \ac{RBMT} systems lies in their complexity of building the models, which is associated with the manual processing of deep linguistic knowledge. For example, Interlingual approaches require significant time to create the internal representation by linguists~\cite{dorr2004machine}. 
Linguistics professionals commonly need to annotate the rules and structure manually. Although \ac{RBMT} are clearly rather expensive approaches, grammar problems may be found easily by linguists. 
Note that the modification of current rules does not guarantee better results. Linguists even may diverge from some rules, thus causing the creation of inconsistent rules. Consequently, the creation of accurate \ac{RBMT} has come to be regarded as a very difficult task with many bottlenecks. Additionally, \ac{RBMT} performs well with closely related languages, however, the performance drops between language pairs of significant difference.

On the other hand, \ac{SMT} systems do not require complex linguistic rules. However, the main challenge is reaching a good translation quality over multi-domain texts. Commonly, \ac{SMT} systems are trained using reference translations by which \ac{ML} algorithms are able to analyze the data and find patterns by themselves, thus being able to translate text without any rules created by humans. Although some basic linguistics mistakes have been solved by \emph{Tree-based} and \emph{Neural Network-based} approaches, the lack of complex linguistic rules still causes ambiguity problems (e.g., errors on re\-la\-ti\-ve pro\-nouns)\-~\cite{bisazza2016survey}. An additional problem of the latter approach is the complexity of performing error analysis over outputs. For instance, \ac{NMT} systems do not provide an easy way to find problems in a feasible time~\cite{johnson2016google,wu2016google}. In spite of the fact that \ac{NMT} approaches have currently been achieving better results than the others, they have similar drawbacks and are much less interpretable than \ac{SMT}. Additionally, \ac{NMT} approaches struggles to deal with \ac{OOV} words (rare words) since they have a fixed vocabulary size. However, the community has been combining  efforts in order to handle this problem by using character-based models or sub-words units, which are able to predict unforeseen words\cite{sennrich2015neural,luong2016achieving,chung2016character}. However, it is still an open problem which is related to the disambiguation of words. Furthermore, despite the advances in \ac{NMT} approaches, a significant amount of effort is being invested in improving the scalability of \ac{RBMT} approaches in order to achieve a higher performance than their counterparts. Hence, \ac{RBMT} is still regarded as essential to high-quality translation, even more so for rich morphological languages.

\subsection{Open MT Challenges}
\label{sec:openMTproblems}

The most problematic unresolved \ac{MT} challenges, from our point of view, which are still experienced by the aforementioned \ac{MT} approaches are the following:

\begin{enumerate}

\item  \emph{Complex semantic ambiguity}: This challenge is mostly caused by the existence of ho\-mo\-ny\-mous and po\-ly\-se\-mous words. Given that a significant amount of parallel data is necessary to translate such words and expressions adequately. \ac{MT} systems commonly struggle to translate these words correctly, even if the models are built upon from 5- or 7-grams. For exam\-ple, ``John \underline{promises to keep} his room tidy" and ``John has some \underline{promises to keep} until he is trusted again". Although the meaning of both are clear to humans, these sentences for \ac{SMT} systems are statistically expensive and prone to failure\footnote{See a complete discussion about the problem:~\url{http://tinyurl.com/yck5ngj8}}. Additionally, for translating the simple word ``bank", context information is essential for determining which meaning to assign to it. 

\item  \emph{Structural divergence}: By definition, structural reordering is reorganizing the order of the syntactic constituents of a language according to its original structure~\cite{bisazza2016survey}.
It in turn becomes a critical issue because it is the core of the translation process. Every language has its own syntax, thus each \ac{MT} system needs to have adequate models for the syntax of each language. For instance, reordering a sentence from Japanese to English is one of the most challenging techniques because of the SVO (sub\-ject-verb-ob\-je\-ct) and SOV (sub\-ject-object-verb) word-order  difference and also, one English word often groups multiple meanings of Japanese characters. For example, Japanese characters make subtle distinctions between homonyms that would not be clear in a phonetic language such as English.

\item  \emph{Linguistic properties/features}: A large number of languages display a complex tense system. When con\-fron\-ted with sentences from such languages, it can be hard for \ac{MT} systems to recognize the current input tense and to translate the input sentence into the right tense in the target language. For instance, some irregular verbs in English like ``set'' and ``put'' cannot be determined to be in the present or past tense without previous knowledge or pre-processing techniques when translated to morphologically rich languages, e.g., Portuguese, German or Slavic languages. Additionally, the grammatical gender of words in such morphologically rich languages contributes to the problem of tense generation where a certain \ac{MT} system has to decide which inflection to use for a given word. This challenge is a direct consequence of the structural reordering issue and remains a significant problem for modern translator systems.
\end{enumerate}

Additionally, there are five \ac{MT} open challenges posed by Lopez and Post~\cite{lopez2013beyond} which we describe more generically below. 

(1) Excessive focus on English and European languages as one of the involved languages in \ac{MT} approaches and poor research on low-resource language pairs such as African and/or South American languages. (2) The limitations of \ac{SMT} approa\-ches for translating across domains. Most \ac{MT} systems exhibit good performance on law and the legislative domains due to the large amount of data provided by the European Union. In contrast, translations performed on sports and life-hacks commonly fail, because of the lack of training data. (3) How to translate the huge amount of data from social networks that uniquely deal with no-standard speech texts from users (e.g., tweets). (4) The difficult translations among morphologically rich languages. This challenge shares the same problem with the first one, namely that most research work focuses on English as one of the involved languages. Therefore, \ac{MT} systems which translate content between, for instance, Arabic and Spanish are rare. (5) For the speech translation task, the parallel data for training differs widely from real user speech.

The challenges above are clearly not independent, which means that addressing one of them can have an impact on the others. Since \ac{NMT} has shown impressive results on reordering, the main problem turns out to be the disambiguation process (both syntactically and semantically) in \ac{SMT} approaches~\cite{Koehn2010}.

\subsection{\ac{MT} Evaluation Metrics}

An evaluation of a given \ac{MT} system may be carried out either automatically or manually. Generally, the \ac{MT} community has opted to use automatic metrics to decrease human efforts and time. Once an \ac{MT} system has shown promising results from a certain automatic evaluation metric, the output of this \ac{MT} system is then evaluated by a human translator. A common process of automatic evaluation is composed of the source text (input), target text (output produced by an \ac{MT} system which is also called hypothesis) and the reference translation of the source text (output created by a human translator). The reference translation is compared automatically against the target text using a given evaluation metric. There are plenty of automatic \ac{MT} evaluation metrics. However, below we introduce only the most important metrics which were also used for evaluating the surveyed papers. Such metrics enable a scientific comparison between the quality of different \ac{MT} systems.

\begin{itemize}

    \item \textbf{BLEU} was created by Papineni~\cite{papineni2002bleu} as an attempt to decrease the professional translation efforts for evaluating the performance of \ac{MT} systems. BLEU is widely chosen for evaluating \ac{MT} outputs due to its low costs. BLEU uses a modified precision metric for comparing the \ac{MT} output with the reference translation. The precision is calculated by measuring the ngram similarity (size 1-4) at word levels. BLEU also applies a brevity penalty by comparing the length of the \ac{MT} output with the reference translation. Additionally, some BLEU variations have been proposed to improve its evaluation quality. The most common variation deals with the number variability (frequency) of useless words commonly generated by \ac{MT} systems~\cite{babych2004extending}. However, the main weakness of BLEU is its difficulty handling semantic variations (i.e., synonyms) while performing the ngram similarity.

    \item \textbf{NIST} was designed by Doddington~\cite{doddington2002automatic} to address some weaknesses of BLEU, upon which it is based. Instead of attributing the same weight for each word in a sentence, NIST gives more weight to a rare word when match\-ed by n-gram. It also modifies the penalty applied by BLEU on the comparison of the \ac{MT} output with the human translation, reducing the impact of small variations on the overall score. 

    \item \textbf{METEOR} was introduced by Banerjee and La\-vie~\cite{banerjee2005meteor} to overcome some weaknesses of BLEU and NIST, for example, the lack of explicit word-matching between translation and reference, the lack of recall and the use of geometric averaging of n-grams. The goal of METEOR is to use semantic features to improve  correlation with human judgments of translation quality. To this end, METEOR 
    considers the synonymy overlap through a shared WordNet synset of the words.

    \item \textbf{TER} is different from the aforementioned metrics. TER measures the number of necessary edits in an \ac{MT} output to match the reference translation exactly. 
    The goal of TER is to measure how much effort is needed to fix an automated translation to make it fluent and correct~\cite{snover06astudy}. The edits consist of insertions, deletions, substitutions and shift of words, as well as capitalization and punctuation. The TER score is calculated by computing the number of edits divided by the average referenced words.
    
    \item \textbf{MEANT} was firstly created by Chi-kiu Lo in 2011~\cite{lo2011meant} to alleviate the semantic correlation deficit between the reference and \ac{MT} outputs from well-known \ac{MT} metrics such as METEOR and BLEU. Although METEOR has been created to deal with the semantic weakness of BLEU (i.e synonyms), it ignores the meaning structures of the translations. MEANT outperforms all the existing \ac{MT} metrics in correlation with human adequacy judgment, and is relatively easy to port to other languages. MEANT requires only an automatic semantic parser and a monolingual corpus of the output language, which is used to train the discrete context vector model to compute the lexical similarity between the semantic role fillers of the reference and translation. Recently, a variation of MEANT has been created by Chi-kiu Lo et. al~\cite{lo2015improving} which is currently the state-of-the-art of semantic \ac{MT} evaluation metrics. This new version of MEANT repla\-ced the discrete context vector model with continuous word embeddings in order to further improve the accuracy of MEANT. The accuracy of MEANT relies heavily on the accuracy of the model that determines the lexical similarities of the semantic role fillers.
    
    \item \textbf{chrF} proposed by Popovi{\'c}~\cite{popovic2015chrf,popovic2016chrf} was initially the use of character n-gram precision and recall (F-score) for automatic evaluation of  \ac{MT} outputs. Recently, Popo\-vi{\'c}~\cite{popovic2017chrf++} enhanced chrF with word n-grams which has shown appealing results, especially for morphologically rich target languages. Although n-gram is already used in well-known and complex \ac{MT} metrics, the investigation of n-grams as an individual metric has not been exploited before. chrF has shown a good correlation with human rankings of different \ac{MT} outputs. chrF is simple and does not require any additional information. Additionally, chrF is language and tokenisation independent.

\end{itemize}

\section{Surveyed Papers}
\label{sec:mtsw}
In this section, we describe how the surveyed articles work and conclude with suggested directions for using \ac{SWT} in \ac{MT}\footnote{The description of the articles may not follow a standard due to the lack of details in some works.}.

\subsection{Selected Research Works}

This section describes all surveyed articles according to the \ac{MT} approaches described in~\autoref{sec:mtapproaches} in order to provide continuity for the reader. \autoref{directly} presents a comparison of all surveyed articles, along with their \ac{MT} method and applied \ac{SWT}. The column \textit{MT approach} represents which kind of \ac{MT} was chosen for handling the translation process. The column \textit{SW method} denotes what \ac{SWT} were used to extract the knowledge contained in a given \ac{SW} resource, for instance, semantic annotation technique, SPARQL queries, and reasoning. Semantic annotation is the process of inserting additional information or metadata to  concepts in a given text or any other content. It enriches the content with machine-readable information. SPARQL is a \ac{RDF} query language, which is a semantic query language for databases. Reasoning is the technique responsible for deriving unseen logical and sensible relations from a set of explicit facts or axioms. The column \textit{Resource} shows what kind of resource was used to acquire the \ac{SW} knowledge, in this case ontology files or \ac{LOD}, which means a \ac{KB} such as DBpedia or BabelNet. The column \textit{Evaluation} illustrates the evaluation process that was applied for measuring the quality of the work, e.g., whether the evaluation was performed by humans or using automatic evaluation metrics. 

\begin{table*}[htb]
\centering
\caption{Details of surveyed articles}
\label{directly}
\footnotesize
\begin{tabular}{@{} l|c|c|c|c|c @{}}
\hline
\textbf{Citation} & \textbf{Year} & \textbf{MT approach}  & \textbf{SW method} & \textbf{SW resource} & \textbf{Evaluation}\\
\hline
C. Vertan~\cite{vertan2004language} & 2004 & EBMT & Annotation & Ontologies & None\\
\hline
N. Elita and A. Birladeanu~\cite{elita2005first} & 2005 & EBMT & SPARQL & Ontologies & None\\
\hline
W. Hahn and C. Vertan~\cite{hahn2005challenges} & 2005 & EBMT & SPARQL + Annotation & Ontologies & None\\
\hline
C. Shi and H. Wang~\cite{shi2005research} & 2005 & None & Reasoner & Ontologies & None\\
\hline
N. Elita and M. Gavrila~\cite{elita2006enhancing} & 2006 & EBMT & SPARQL & Ontologies  & Human\\
\hline
E. Seo et al.~\cite{seo2009syntactic} & 2009 & None & Reasoner & Ontologies  & None\\
\hline
P. Knoth et al.~\cite{knoth2010facilitating} & 2010 & RBMT or SMT & Annotation & Ontologies & Human \\
\hline
A. M. Almasoud and H. S. Al-Khalifa~\cite{almasoud2011proposed} & 2011 & TBMT + EBMT & SPARQL & Ontologies & Human\\
\hline
L. Lesmo et al.~\cite{lesmo2011ontology} & 2011 & Interlingua & Annotation & Ontologies & Human\\
\hline
B. Harriehausen-M\"uhlbauer and T. Heuss~\cite{harriehausen2012semantic,heuss2013lessons} & 2012 & Direct & SPARQL + Reasoner & Ontologies & Human\\
\hline
K. Nebhi et al.~\cite{nebhi2013nerits} & 2013 & TBMT & Annotation & LOD  & None\\
\hline
J. P. McCrae and P. Cimiano~\cite{mccrae2013mining} & 2013 & SMT & Annotation & LOD & Human\\
\hline
D. Moussallem and R. Choren~\cite{moussallem2015using} & 2015 & SMT & SPARQL & Ontologies & Human \\
\hline
O. Lozynska and M. Davydov~\cite{lozynska2015information} & 2015 & RBMT & Annotation & Ontologies & Human \\
\hline
K.Simov et al.~\cite{simovtowards} & 2016 & RBMT + SMT & SPARQL & LOD & Automatic\\
\hline
T.S. Santosh Kumar.~\cite{kumar2016} & 2016 & SMT & SPARQL & Ontologies & Human\\
\hline
N. Abdulaziz et al.~\cite{neama2016towards} & 2016 & SMT & SPARQL & Ontologies & Human\\
\hline
J. Du et al.~\cite{jinhua2016} & 2016 & SMT & SPARQL & LOD & Automatic\\
\hline
A. Srivastava et al.~\cite{srivastava2016} & 2016 & SMT & SPARQL + Annotation & LOD & Automatic\\
\hline
C. Shi et al.~\cite{shi2016knowledge} & 2016 & NMT & Annotation & LOD & Automatic + Human \\
\hline
A. Srivastava et al.~\cite{srivastava2017} & 2017 & SMT & SPARQL + Annotation & LOD & Automatic\\
\bottomrule
\end{tabular}
\end{table*}

\subsubsection{Direct}

\begin{itemize}
    \item Heuss et~al.\cite{harriehausen2012semantic} combined \ac{SWT} as a post-editing technique with the direct approach in their work. 
Post-editing technique involves fixing mistakes from a given output by choosing the right translated word or order. Although this technique is commonly used by a professional translator or a linguist, its automated implementation has been widely researched recently in order to reduce human efforts. 
The works~\cite{harriehausen2012semantic,heuss2013lessons} propose a method for retrieving translations from a domain ontology. The approach performs SPARQL queries to search  translations of a given word. The ontology uses SKOS vocabulary~\cite{Miles2005} to describe its multilingual content. SKOS is a common model for describing concepts in \ac{SW}. It uses the \emph{prefLabel} property to assign a primary label to a particular concept and \emph{altLabel} for alternative names or translations\footnote{\url{https://www.w3.org/TR/skos-reference/}}. Once SPARQL queries have retrieved the translations, a reasoner infers a relationship between the source and target words based on its properties contained in the ontology. The authors manually marked parts of the data as triggers. 
Hence, the inference rules were basically done manually. Although this work provides important insights, the evaluation of its output is insubstantial as the authors evaluated only one sentence to validate their approach. The given explanation for this weak evaluation is that the reasoner is not fast enough for large texts and for executing on more than one sentence. 


\end{itemize}

\subsubsection{Transfer-Based}

\begin{itemize}
    \item In NERITS~\cite{nebhi2013nerits}, the authors presented an \ac{MT} system with additional information about the translated text. They used DBpedia to provide concepts about each word, similar to the semantic annotation technique. A semantic annotation technique involves annotating a given word or text, adding information/concepts about it from a given \ac{KB}. The goal of NERITS is to provide more knowledge for users using \ac{SW}. The authors chose a \emph{Transfer-based} approach for translating the content. 
    The translation process begins performing \ac{NER}~\cite{nadeau2007survey} through a tool called Wiki\-meta to deal with the variations of entities.
    The \ac{NER} technique is responsible for recognizing the entities and their respective types. For instance, Microsoft is an organization and Bill Gates is a person. Thus, Wikimeta is responsible for linking each recognized entity with a given resource within the knowledge base.
    Although the authors contended for the use of a post-editing technique, the \ac{SW} method applied here does not edit or improve the output, it just provides concepts for helping the user to understand and assimilate the translation. The authors stated that no current \ac{MT} systems provide conceptual knowledge about the translated words, they only present the crude translation. Their goal was to describe how users can learn about languages combining \ac{MT} with \ac{SW}. The authors stated that their proposal could not be evaluated with automatic \ac{MT} metrics because they did not improve the translation process, but annotated the translations providing additional knowledge. 
    
    \item O. Lozynska and M. Davydov~\cite{lozynska2015information} developed an \ac{MT} system for translation of \ac{USL}. The authors chose the Transfer-based \ac{MT} approach because of the lack of parallel data for statistical translation. Additionally, the authors argued that a rule-based approach, along with ontologies, is best suited for implementing the rules of sign languages. To this end, the authors used a grammatically augmented ontology from a variety of domains such as education, nature and army. The ontology was mainly used for supporting a given parser in the extraction of syntactic and semantic rules. These syntactic-semantic rules enable a deep analysis of sentences thus avoiding the problem of ambiguity in \ac{USL}. The crucial contribution according to authors was the creation of a new domain-specific language based on the ontologies which may be further used for editing and processing future works in the translation of \ac{USL} using ontologies. The evaluation was carried  out manually by linguists since the sign languages are difficult to be evaluated automatically. The results were quite promising, as the ontology contributed to 19\% of improvements in the analysis of Ukrainian signs and 35\% in the generation of Ukrainian natural language compared to baseline from related works.   
  
  \end{itemize}
\subsubsection{Interlingua} \label{sec:interlingua}

\begin{itemize}
    \item Lesmo et~al.~\cite{lesmo2011ontology} presented an approach for translating from the Italian language to Italian Sign Language. The authors translated the content using the interlingual approach with \ac{SWT}. The goal of using \ac{SWT}, in this case an ontology, is to discern the syntactic ambiguity of a given word caused by the automatic syntax-semantic interpretation within the generation step at the translation process. To this end, the authors used a very limited domain, a weather forecast ontology, for dealing with the variety of specific terms. Therefore, the authors assumed that a certain word meaning may be expressed by ontology nodes. Thus, the relationship between nodes which are ontology properties may resolve the syntax-semantic issues when translating a sentence. Afterwards, the authors converted the ontology structure into first-order logic and included the logic forms using the OpenCCG tool into the interlingual language model. Therefore, the translation starts by performing a syntactic analysis step which is based on dependency trees. Afterward, each word is annotated with its respective lexical meaning from the ontology, thus creating an ontological representation. This representation is then interpreted to determine which path (i.e., meaning) to follow and then translating the words correctly. The Interlingual approach was chosen because a translation of sign language requires deep linguistic expertise. Finally, the authors raised the problem of the lengthy time taken to translate due to the many ambiguity issues related to sign language and concluded that the approach requires further evaluation.  
    
\end{itemize}

\subsubsection{Statistical}

\begin{itemize}
    

\item J. P. McCrae and P. Cimiano~\cite{mccrae2013mining} extracted English-Ger\-man translations from \ac{LOD}. They used these results to improve the \ac{WSD} inside \ac{SMT}. The authors used the well-known \ac{SMT} system called Moses~\cite{koehn2007moses}, inserting the results retrieved from \ac{LOD} into the phrase table created by Moses. The phrase table, which contains the frequency of the phrases, is used to select the best meaning (i.e., translation), thus discerning among various source meanings to arrive at a unique target meaning. Each word mined from \ac{LOD} is queried in the Moses frequency table. If the word is not in the table, they insert the best-fit word (1.0) into the table. They did not achieve good results with automatic evaluation using BLEU~\cite{papineni2002bleu}. The outcomes showed no improvement against Moses as a baseline system. Baseline achieved 11.80 while baseline+ \-\ac{LD} achieved 11.78. However, the manual evaluation done by linguists showed a significant improvement of almost 50\% percent in the output translation.

\item Moussallem and Choren~\cite{moussallem2015using} presented a novel ap\-proach to tackle the ambiguity gap of message translations in dialogue systems. The translations occurred between Brazilian Portuguese and English. Currently, submitted messages to a dialogue system are considered as isolated sentences. Thus, missing context information impedes the disambiguation of homographs in ambiguous sentences. Their approach tries to solve this disambiguation problem by using concepts in different existing ontologies. First, the user log is parsed in order to find out the respective ontology that matches the dialogue context. Using the SKOS vocabulary, the ontology that returns the most results for each verb or noun is used as context. When the \ac{SMT} system returns a translation indicating a homonymous word (i.e., many different possible translations), this method queries the word in the ontology using SPARQL and replaces the target word accordingly. The authors evaluated their approach manually using empirical methods. The ontologies were Music ontology\footnote{\url{http://musicontology.com/}} and Vehicle ontology\footnote{\url{http://www.heppnetz.de/ontologies/vso/ns}}.
Regarding its focus on dialogue systems, the work fails to translate slang or expressions that are common in dialogues. Since such sentences are rarely in a well-structured form, the translations often contain structural errors. Also, the context is fixed during a dialogue, so a topic change is not reflected by the system\footnote{\url{https://github.com/DiegoMoussallem/judgemethod}}. 

\item Neama Abdulaziz et al.~\cite{neama2016towards} extended the previously described approach presented by Moussallem et~al. to Arabic-English translations. As its main extension, this work includes a dependency tree parser because of the rich morphological structure of Arabic. Thus, the \ac{SMT} approach used by authors is \emph{Tree-based}, which inserts the tree as statistical rules. In contrast, during the common training phase of \ac{SMT}, the dependency rules are attached according to the sent message. The authors use the MS-ATKS tool for analyzing the syntactic structure of messages and also a domain ontology on the Quran. Once this work has been applied to dialogue systems, the evaluation is carried out empirically by humans. The authors propose to apply different automatic metrics as further plans in order to evaluate additional aspects of this work.      

\item Santosh Kumar T.S.~\cite{kumar2016} proposed an approach to address lexical disambiguation in the translation of proper nouns from English to Dravidian languages such as Tamil. To this end, the authors created a corpus containing only ambiguous sentences for testing their approach. They used Google Translate as an \ac{SMT} system because, in accordance with the state-of-the-art, Google \ac{MT} system has the best translation model for translating from English to Tamil. The approach preprocessed ambiguous sentences using an NER tool for recognizing persons. Additionally, they relied on FOAF ontology in RDF structure for supporting the translation. This ontology contains information about a large number of names in Tamil. Due to the lack of resources in the Tamil language, which are required for automatic metrics such as BLEU, the evaluation was carried out manually. It presented positive outcomes where the professional translator confirmed the capability of \ac{SWT} to support translations of entities among morphologically rich languages.   

\item Jinhua Du et~al.~\cite{jinhua2016} created an approach addressing the problem of \ac{OOV} words. These kinds of words do not appear in the translation table of a given \ac{SMT} approach such as \emph{Phrase-based}. Commonly, they are na\-med entities which often appear on the Web, such as persons and organizations, but they can also be common words like ``kiwi" which is highly ambiguous. 
Therefore, the authors proposed three methods to deal with the \ac{OOV} words problem 
using a \ac{SW} resource called BabelNet. The respective methods are: (1) Direct Training, which retrieves every pair of source and target words, creating a dictionary to use during the training phase. (2) Domain-Adaption, which recognizes the subject of source corpus and applies a topic-modelling technique as Moussallem et al. have done. Hence, the found subject adapts the target corpus for a specific domain, gathering the information from BabelNet. According to the authors, these first two methods did not perform well, so they decided to apply BabelNet as (3) Post-processing technique. To this end, they used an \ac{SMT} decoder in order to recognize \ac{OOV} words which were not translated beforehand. Subsequently, they performed SPARQL queries through BabelNet API~\cite{navigli2012multilingual} to retrieve translations of these words. As in previously surveyed works, they used Moses as a baseline for an \ac{SMT} system to perform experiments. Additionally, they chose Chinese, English and Polish languages for translating the contents and decided to evaluate the translations from their system using the automatic metrics BLEU and TER. 
The evaluation showed unstable results using BabelNet. However, it also uncovered a promising way to rectify the \ac{OOV} problem by looking for unknown words in \ac{LD}. The authors intend to investigate different forms of using \ac{SWT} in \ac{MT} systems as a future work.

\item A. Srivastava et al.~\cite{srivastava2016} implemented a novel framework which translates named entities in \ac{SMT} systems using \ac{LOD} resources. The framework is akin to the approach of J. P. McCrae and P. Cimiano~\cite{mccrae2013mining} and it uses Moses as an \ac{SMT} baseline system. \ac{NIF} vocabulary is used as interchanging format for converting natural language sentences to \ac{SW} structure (i.e., triples). Then, the authors used DBpedia Spotlight~\cite{spotlight} to recognize  entities, thus facilitating their translation. Once the entities are recognized, their respective translations are gathered using SPARQL through the DBpedia endpoint. The translation of entities is marked using XML format and inserted into a Moses translation table. These markings avoid Moses having to translate the entities using its own probabilistic model and forces Moses to select the translations retrieved from \ac{LOD} resources. The evaluation was carried out using the IT domain data from the \ac{MT} workshop,\footnote{\url{http://www.statmt.org/wmt16/it-translation-task.html}} which consists of 1.000 sentence translations. 
This task uses BLEU and TER for measuring the performance of a given \ac{MT} system. Their \ac{LD}-based framework showed a significant improvement of 12\% in comparison to the Moses baseline. BLEU increased by 0.8\% and TER decreased by 3.2\%. The authors dubbed their framework ``\ac{SW}-aware \ac{MT} system'. Although they provided links to projects related to this framework, there are no links to their implementation.

\item C. Shi et al~\cite{shi2016knowledge} built a semantic embedding model relying on knowledge-bases to be used in \ac{NMT} systems. The work is dubbed \ac{KBSE}, which consists of mapping a source sentence to triples and then using these triples to extract the internal meaning of words to generate a target sentence. The mapping results in a semantic embedding model containing \ac{KB} triples which were responsible for gathering the key information of each word in the sentences. Therefore, the authors investigated the contribution of \ac{KB} to enhance the quality of the translation of Chinese-English \ac{MT} systems. To this end, they applied \ac{KBSE} in two domain-specific datasets, electric business, and movies. The evaluation was two-fold. First, they compared \ac{KBSE} with a standard \ac{NMT} system and also with Moses as \ac{SMT} system by using BLEU as an automatic evaluation metric. Second, they selected humans to manually evaluate the translations. Additionally, the authors used an external named entity translator to get the translation of English entities into Chinese. Moreover, they included this entity translator in the evaluation of the other \ac{MT} systems for a fair comparison. The results of \ac{KBSE} were quite promising, BLEU showed an improvement of 1.9 points (electric) and 3.6 (movie) when compared to the standard encoder-decoder \ac{NMT} system. Additionally, using the \ac{KBSE} method received much higher results than using the Moses toolkit. This work shows that a given neural model, when trained using semantic information gathered from a \ac{KB}, is able to memorize the key information of source sentences. Also, their model rarely made grammatical mistakes as the authors expected because of the strong learning ability of Gated Recurrent Unit~\cite{cho2014learning}. Moreover, \ac{KBSE} presented some errors when translating entities due to the external entity translator. Furthermore, since \ac{KBSE} comprises two separate models, when an error appears at the source part, it can not be corrected in the target part.

\item A. Srivastava et al.~\cite{srivastava2017} mixed approaches from their last work~\cite{srivastava2016} with some strategies implemented by Jinhua Du et~al.~\cite{jinhua2016} and the approach implemented by J. P. McCrae and P. Cimiano~\cite{mccrae2013mining}. This work therefore comprises of three strategies for using \ac{LD} with Mo\-ses as an \ac{SMT} system. In the first strategy, the authors used the \ac{LD} resources as dictionaries for word alignment. Thus the translation models contained knowledge from the \ac{LD} resources and also from bilingual corpora. Hence, the Moses decoder is able to decide which translation of a given word to choose. This strategy is akin to J. P. McCrae and P. Cimiano's~\cite{mccrae2013mining} work. The second strategy relied on their former work, \cite{srivastava2016} which forced the Moses decoder to retrieve the translation of a given named entity by performing \ac{EL} along with SPARQL queries from \ac{LD}. The last strategy was inspired by Jinhua Du et~al.'s~\cite{jinhua2016} work. It applied a post-editing technique to correct the \ac{OOV} words, which means untranslated words in \ac{MT} output. The authors evaluated their approach using BabelNet, DBpedia and JRC names as \ac{LOD} resources on the WMT12 dataset\footnote{\url{http://www.statmt.org/wmt12/}}. The evaluation showed only modest improvements by BLEU and TER metrics because both deal poorly with semantics, i.e, synonyms. Therefore, the real contribution of \ac{LD} in this work was made clear when the evaluation was performed by humans.  



\end{itemize}

\subsubsection{Example-Based}

\ac{EBMT} was the first methodology proposed to work alongside \ac{SWT} because of its architecture. The \ac{EBMT} process is simpler than other methods, thus facilitating its combination with other technologies. All works discussed in this section were carried out by a single research group. However, the research was discontinued. 

\begin{itemize}
     
\item Vertan~\cite{vertan2004language} presented an architecture based on \ac{EBMT} which retrieves word meanings using semantic annotation from ontologies. As \ac{SW} was in its infancy in 2004 when this work was undertaken, it proposed an architecture but did not implement it or give details properly. Thus, it is unclear how the semantic annotation actually helped the translation process. 

\item Her second work with Hahn ~\cite{hahn2005challenges} showed how semantic annotations could be used in \ac{MT} systems. They explained how the properties in \ac{RDF} data structures could support the annotation of words in several languages. Vertan and Hahn presented the same architecture of Vertan's previous work~\cite{vertan2004language}, but did not show the process of evaluation.  

\item A year later, Elita et~al.~\cite{elita2005first} proposed a prototype of \ac{EBMT} where \ac{SW} was more concrete, which queried sentences from ontologies. The sentences were extracted from the text and queried through SPARQL endpoints instead of common databases. The translation was only provided by the ontology, which retrieved exactly one translation for each source sentence. Consequently, this work showed how Vertan's architecture could be implemented~\cite{vertan2004language}. However, no evaluation was provided.

\item Subsequently, N. Elita and M. Gavrila~\cite{elita2006enhancing} used an ontology to map existing concepts in a source text using \ac{NER} and summarization techniques. The authors used five languages to prove their method. This approach showed how an ontology can support an automatic \ac{WSD} process in \ac{MT}. The authors stated that there was no need to analyze the text syntactically. However, this lack of analysis prevented the approach from handling inflections properly, which is a requirement for achieving good translations and thus a negative point of this work. The evaluation was done manually and demonstrated solid improvements only on translation templates, which their approach identified beforehand, but failed to present a comparison to other \emph{example-based} approaches.
\end{itemize}

\subsubsection{Hybrid}

Whilst the combination of multiple \ac{MT} approaches is already a challenge, three works were found that integrate a hybrid approach with \ac{SWT}. 

\begin{itemize}

    \item Almasoud et~al.~\cite{almasoud2011proposed} mix the translation process of \ac{TBMT} (a rule-based approach) with \ac{EBMT} (a corpus-based one) while using a domain ontology. This work translates from the Arabic language to Arabic Sign language. According to the authors, \ac{TBMT} was chosen because a sign language requires deep linguistic knowledge (like~\cite{lesmo2011ontology} did in~\autoref{sec:interlingua}). First, the text is converted into a representation model based on Arabic sign language rules, then the model is translated into signals. The ontology which belongs to the religion domain is applied in a similar fashion to the \ac{WSD} task, exploiting its semantic knowledge. Then, every word is searched in the ontology to find its respective sign. In the case where there is no sign for a certain word, the SignWriting corpus is used to retrieve respective synonyms. Like many other works in this survey, this work fails to provide a detailed evaluation. The authors argued that an evaluation with automatic metrics is not possible because these do not provide measures to evaluate sign languages. Thus, they performed a manual evaluation, but did not provide any measure of improvement.
    
    \item Knoth~et al.~\cite{knoth2010facilitating} presented an approach combining \ac{CLIR} and \ac{MT} ap\-proaches to gather translations of specific terminologies from domain ontologies. \ac{CLIR} is a technique for retrieving contents from languages other than the language used for searching. They did not mention which kind of \ac{RBMT} and \ac{SMT} method they combined. Their approach is divided into two phases: (1) The initialization phase is responsible for creating the monolingual ontology and generating the lightweight multilingual ontology using a given \ac{RBMT} system. To do this they first collected the monolingual text from the European Government and built a simple ontology using the SKOS vocabulary. Using \ac{RBMT}, they translated the monolingual terms into nine European languages, thus creating a multilingual lightweight ontology. This multilingual ontology is then evaluated by domain experts, concluding the first step. (2) The bootstrapping phase then applied \ac{CLIR} techniques by querying terms through SPARQL. Multilingual terms were used to create parallel corpora for each of the nine languages in order to train the \ac{SMT}. Regarding the parallel corpora, they extracted translation pairs and updated the terms when a new document was submitted. Their approach allows the user to decide which domain and language to use. The authors emphasized that the \ac{CLIR} method can be adopted by any \ac{MT} system. Furthermore, they stressed that multilingual ontologies support translations between any pair of languages. They also did not perform a proper evaluation of their architecture.
    
    \item Simov et al.~\cite{simovtowards} aimed to create a semantic-based hybrid \ac{MT} system between Bulgarian and English in the domain of information technology. Their system supports the automatic identification of appropriate answers to user questions in a multilingual question/answering system. The authors chose \ac{RBMT} because Bulgarian is a morphologically rich language. However, they do not mention which rule-based approach was used. Along with the \ac{RBMT}, they used dictionaries and a PoS tagger in order to create rules which were then included in the \ac{SMT} model akin to a \emph{tree-based} approach. Consequently, the \ac{SMT} system is the main part of their \ac{MT} system. They used a parallel corpus as a gold standard dataset from EUROPARL\footnote{\url{http://www.statmt.org/europarl/}} to evaluate their approach. Additionally, they used two \ac{WSD} techniques which were supported by OntoWordNet and LT4eL~\cite{monachesi2006language}. Although they used LT4eL as a domain ontology, they suggested that DBpedia may be used to address other domains, because DBpedia fails in covering the information technology domain. 
    LT4el was included in OntoWordNet which is their previous work~\cite{gangemi2003ontowordnet}. Thus, they were able to support \ac{WSD} through the WordNet synsets inside Moses \ac{SMT} system. The evaluation, comparing their hybrid \ac{MT} system with the Moses system as baseline, showed no improvement for the BLEU metric. However, when the authors evaluated using the NIST metric, their system showed real improvement. The authors propose to generate more sophisticated rules for further improvements.
\end{itemize}

\subsubsection{New perspective -- Ontology-based \ac{MT}}
\label{sec:newparadigm}

In this section, we describe two articles using \ac{SWT} to translate contents without relying on well-known \ac{MT} approaches. They present an \ac{MT} system mainly based on \ac{SW} methods.

\begin{itemize}

\item In the first work, Seo et~al.~\cite{seo2009syntactic} use an ontology created by themselves from a gold standard corpus by performing the ontology learning method~\cite{buitelaar2005ontology}. This ontology is responsible for providing semantic knowledge of words. The syntactic analysis uses formal language expressed using \ac{EBNF} notation~\cite{essalmi2006graphical} to match the English patterns defined beforehand. These patterns are only defined in the English language, hence the \ac{MT} system only translates contents from English to Korean. Their system is not able to perform a translation in both directions (round-trip translations). Co-occurring words in the corpus are represented by relationship properties in the ontology. 

Thus, the algorithm performs \ac{WSD} by analyzing the ontological relationships between part-of-speech tags for deciding the best translation. For instance, the Korean language is well-known to group meaning of several words together, like the Chinese language. Thus, when translating a given word, its  meaning has to be determined. In the example they gave, they attributed the translation of ``hard" to a given Korean character according to the part-of-speech given by the \ac{EBNF} notation. Then, they were able to search for a translation of ``hard" as an adverb having the meaning ``with effort". This method is comparable to a bag-of-words algorithm, but with additional graph features provided by the ontology to deal with Korean language structure.

\item The second work by Shi et~al.~\cite{shi2005research} proposes an ontology they also created. The ontology ``SCIENTIST" describes terms in the physical science domain and was derived from other ontologies like SUMO. It is used to improve the semantic quality in their Chinese-English \ac{MT}. The syntactic analysis part is done using Lexical Functional Grammar~\cite{kaplan1989formal}. For the semantic analysis, their \ac{WSD} is achieved by comparing relationship weighting between the parts-of-speech of the words in the ontology, similar to the previous work.

\end{itemize}

Both articles show promising approaches. Even though these works are quite similar, the second is much more detailed and its concepts are more applicable in future works and for supporting other languages. Moreover, unlike the first, the second work is capable of making a round translation. Neither work includes an evaluation, and comparing their results directly is impossible due to their different domain ontologies. Consequently, they are not capable of translating content across contexts. Unfortunately, their source codes are also not publicly available.

\subsection{Suggestions and Possible Directions}

According to the surveyed papers, \ac{SWT} have mostly been applied at the semantic analysis step, rather than at the other stages of the translation process, due to their ability to deal with concepts behind the words and provide knowledge about them. \autoref{overviewSW} presents an overview of the surveyed papers, regarding each generic step of the \ac{MT} approaches supported by \ac{SWT}.

\begin{table*}[htb]
\centering
\caption{Semantic Web technologies in Machine Translation steps}
\label{overviewSW}
\small
\begin{tabular}{|p{2.1cm}|p{2cm}|p{2cm}|p{2cm}|p{2cm}|p{2cm}|p{2cm}|}
 \hline
\multirow{3}{*}{MT approaches} & \multicolumn{6}{c|}{\ac{MT} Architecture} \\
\cline{2-7}
 & \multicolumn{3}{c|}{Analysis} &  \multicolumn{3}{c|}{Generation} \\
\cline{2-7}
 & Morphological& Syntactic & Semantic& Morphological & Syntactic & Semantic \\

 \hline
 Direct  &  X & & &  X & & \\
 \hline
 Transfer-based  & &  X & & &  X &  X \\
 \hline
 Interlingua     &   &  X &  X & &  X & X \\
 \hline
Statistical      &  & & X &  &  & X \\
\hline
Example-based    &  X &  &  X & & & X \\
 \hline
Hybrid & X &  X &  X &  X & & \\
 \hline
\end{tabular}
\end{table*}

As \ac{SWT} have developed, they have increasingly been able to resolve some of the open challenges of \ac{MT} described in \autoref{sec:openMTproblems}. They may be applied in different ways according to each \ac{MT} approach. Although the potential graph structure contained in ontologies may act as a disambiguation method with high decision power, some \ac{SW} concepts, such as the alignment of multilingual ontologies and the linking among monolingual knowledge bases,~\cite{gracia2012challenges,labra2015multilingual} still need to be improved before \ac{SWT} can be applied successfully in \ac{MT} systems. 

Additionally, the \ac{SW} community has worked out basic suggestions for generating structured data. However, there are no well-defined rules for building the \ac{KB}s, only common best practices according to Zaveri et al.~\cite{zaveri2015quality}. Due to this lack of defined standards, some research works have produced erroneous ontologies or knowledge graphs into \ac{LOD} repositories. Common issues include the following mistakes: 
\begin{itemize}
\label{commonissues}
    \item Lack of well-defined object properties, e.g., cardinality or reflexiveness.
    \item Mixing of concepts among thesaurus, vocabulary, and ontology.
    \item Incorrect domain and range definitions.
    \item Use of ambiguous annotations.
\end{itemize}

Therefore, applying \ac{SWT} to \ac{MT} may seem difficult. However, recently some efforts from the \ac{LLOD}\footnote{\url{http://linguistics.okfn.org/}} community have been directed to modeling linguistic phenomena, linguistic rules and translations in \ac{LOD} \cite{bosque2015applying}.

In the following subsections, we discuss and suggest future potential solutions to \ac{MT} challenges by applying \ac{SWT}. From our point of view, \ac{SWT} mostly contribute to addressing syntactic and semantic ambiguity problems in \ac{MT} systems. Thus, we divide our suggestions into four categories according to their respective problem.

\subsubsection{Disambiguation}

Human language is very ambiguous. Most words have multiple interpretations depending on the context in which they are mentioned. In the \ac{MT} field, \ac{WSD} techniques are concerned with finding the respective meaning and correct translation to these ambiguous words in target languages. This ambiguity problem was identified early in \ac{MT} development. In 1960 Bar-Hillel~\cite{Bar-Hillel1960} stated that an \ac{MT} system is not able to find the right meaning without a specific knowledge. Although the ambiguity problem has been lessened significantly since the contribution of Carpuat and subsequent works~\cite{carpuat2007improving,navigli2009word,costa2014statistical}, this problem still remains a challenge. As seen in \autoref{sec:mtapproaches}, \ac{MT} systems still try to resolve this problem by using domain specific language models to prefer domain specific expressions, but when translating a highly ambiguous sentence or a short text which covers multiple domains, the languages models are not enough.  

According to surveyed articles, \ac{SW} has already shown its capability for semantic disambiguation of po\-ly\-se\-mous and ho\-mo\-ny\-mous words. However, \ac{SWT} were applied in two ways to support the semantic disambiguation in \ac{MT}. First, the ambiguous words were recognized in the source text before carrying out the translation, applying a pre-editing technique. Second, \ac{SWT} were applied to the output translation in the target language as a post-editing technique. Although applying one of these techniques has increased the quality of a translation, both techniques are tedious to implement when they have to translate common words instead of named entities, then be applied several times to achieve a successful translation. 
The real benefit of \ac{SW} comes from its capacity to provide unseen knowledge about emergent data, which appears every day. Therefore, we suggest performing the topic-modelling technique over the source text to provide a necessary context before translation. Instead of applying the topic-modeling over the entire text, we would follow the principle of communication (i.e from 3 to 5 sentences for describing an idea\cite{prasad2003principles}) and define a context for each piece of text. Thus, at the execution of a translation model in a given \ac{SMT}, we would focus on every word which may be a homonymous or polysemous word. For every word which has more than one translation, a SPARQL query would be required to find the best combination in the current context. Thus, at the translation table, the disambiguation algorithm could search for an appropriate word using different \ac{SW} resources, such as BabelNet\cite{navigli2012babelnet} or DBpedia, in consideration of the context provided by the topic modelling. The goal is to exploit the use of more than one \ac{SW} resource at once for improving the translation of ambiguous terms. The use of two or more \ac{SW} resources simultaneously has not yet been investigated.    

On the other hand, there is also a syntactic disambiguation problem which as yet lacks good solutions. For instance, the English language contains irregular verbs like ``set'' or ``put''. Depending on the structure of a sentence, it is not possible to recognize their verbal tense, e.g., present or past tense. Even statistical approaches trained on huge corpora may fail to find the exact meaning of some words due to the structure of the language. Although this challenge has successfully been dealt with since \ac{NMT} has been used for European languages~\cite{bojar2017findings}, implementations of \ac{NMT} for some non-European languages have not been fully exploited (e.g., Brazilian Portuguese, Latin-America Spanish, Hindi) due to the lack of large bilingual data sets on the Web to be trained on. Thus, we suggest gathering relationships among properties within an ontology by using the reasoning technique for handling this issue. For instance, the sentence ``Anna usually put her notebook on the table for studying" may be annotated using a certain vocabulary and represented by triples. Thus, the verb ``put", which is represented by a predicate that groups essential information about the verbal tense, may support the generation step of a given \ac{MT} system. This sentence usually fails when translated to rich morphological languages, such as Brazilian-Portuguese and Arabic, for which the verb influences the translation of ``usually" to the past tense. In this case, a reasoning technique may support the problem of finding a certain rule behind relationships between source and target texts in the alignment phase (training phase). 

Reasoning techniques have been identified as a possible future way of supporting \ac{MT} tasks because some remaining \ac{MT} issues cannot be solved with explicit knowledge only. For instance, in accordance with Manning~\cite{Manning2011}, some syntactic ambiguity gaps need to be addressed by human knowledge. This kind of human knowledge can be gained using reasoning over ontological relations described by humans. For example, in 2004 Legrand and Pulido~\cite{legrand2004hybrid} combined ontological knowledge with neural algorithms to perform the \ac{WSD} task. The result was very promising, but unfortunately the work was discontinued. Additionally, some researchers, including ~Harriehausen-M\"uhlbauer and Heuss and Seo et~al.~\cite{harriehausen2012semantic, seo2009syntactic}, have used reasoners to disambiguate words in \ac{MT} systems.

For a given reasoning technique to be required, some previous steps need to be addressed. Currently, the Ontology-Lexica Community Group\footnote{\url{https://www.w3.org/community/ontolex/}} at W3C has combined efforts to represent lexical entries, with their linguistic information, in ontologies across languages. Modeling different languages using the same model may provide an alignment between the languages, where it is possible to infer new rules using the language dependency graph structure and visualize a similarity among languages.~\footnote{This insight is already supported by a recent publication in Cicling conference~\url{https://www.cicling.org/2017/posters.html} named ``The Fix-point of Dependency Graph -- A Case Study of Chinese-German Similarity" by Tiansi Dong et al.} \ac{SW} tools that perform reasoning and infer unseen concepts are called Reasoners. Examples include Pellet, RACER, FACT++, and DL-Learner~\cite{abburu2012survey}. For the sake of clarification, DL-Learner~\cite{lehmann2009dl} would learn concepts about a specific source or target text modeled by Ontolex. Thus, DL-Learner could infer new facts across related languages.

Moreover, there are plenty of other ontologies and vocabularies which may support the description of languages, such as Olia~\cite{chiarcos2015olia}, Lemon~\cite{mccrae2011linking}, GOLD~\cite{farrar2003linguistic} and NIF~\cite{hellmann2013integrating}. However, the aforementioned ontology issues may limit reasoner in its ability to support translations. Besides this drawback, a well-known problem of reasoners is the poor run-time performance. Therefore, they are not suitable for real-time \ac{MT} approaches at this point in time. Furthermore, both deficiencies need to be addressed or minimized before implementing reasoners successfully into \ac{MT} systems.     

\subsubsection{Named Entities}

Most \ac{NERD} approaches link recognized entities with database entries or websites~\cite{nadeau2007survey}. This method helps to categorize and summarize text, but also contributes to the disambiguation of words in texts. The primary issue in \ac{MT} systems is caused by common words from a source language that are used as proper nouns in a target language. For instance, the word ``Kiwi" is a family name in New Zealand which comes from the M\=aori culture, but it also can be a fruit, a bird, or a computer program. Named Entities are a common and difficult problem in both \ac{MT} (see Koehn~\cite{Koehn2010}) and \ac{SW} fields. The \ac{SW} achieved important advances in \ac{NERD} using structured data and semantic annotations, e.g., by adding an \texttt{rdf:type} statement which identifies whether a certain kiwi is a fruit~\cite{moussallem2017mag,speck2017ensemble}. In \ac{MT} systems, however, this problem is directly related to the ambiguity problem and therefore has to be resolved in that wider context.

Although \ac{MT} systems include good recognition methods, \ac{NERD} techniques still need improvement. When an \ac{MT} system does not recognize an entity, the translation output often has poor quality, immediately deteriorating the target text readability. 
Therefore, we suggest recognizing such entities before the translation process and first linking them to a reference knowledge base. Afterwards, the type of entities would be agglutinated along with their labels and their translations from a reference knowledge base. For instance, in \ac{NMT}, the idea is to include in the training set for the aforementioned word ``Kiwi", ``Ki\-wi.ani\-mal.link, Ki\-wi.per\-son.link, Ki\-wi.fo\-od.link" then finally to align them with the translations in the target text. In \ac{SMT}, the additional information would be included by \ac{XML} 
or by an additional model. 
This method would also contribute to \ac{OOV} mistakes regarding names. This idea is akin to \citep{shi2016knowledge} where the authors encoded the types of entities along with the words to improve the translation of sentences between Chinese-English. 

Another potential solution could apply type extraction or co-occurrence techniques and combine them with \ac{NERD} methods. The solution would identify which pronouns (e.g., ``it'') are related to entities that have already been mentioned in previous sentences. This would also improve the fluency of target texts. In addition to our suggestions, the contextual linking over texts can help users to acquire a deeper understanding of the translated content (as NERITS has done). Plenty of \ac{SW} tools would be able to support this approach in accordance with Gangemi~\cite{gangemi2013comparison}. This work provides a complete comparison of \ac{SW} tools, detailing their powerful capabilities for extracting knowledge.

\subsubsection{Non-standard speech}

The non-standard language problem is a rather important one in the \ac{MT} field. Many people use the colloquial form to speak and write to each other on social networks. Thus, when \ac{MT} systems are applied on this context, the input text frequently contains slang, \ac{MWE}, and unreasonable abbreviations such as
``Idr = I don't remember.'' and ``cya = see you''. Additionally, idioms contribute to this problem, decreasing the translation quality. Idioms often have an entirely different meaning than their separated word meanings. Consequently, most translation outputs of such expressions contain errors. 

For a good translation, the \ac{MT} system needs to recognize such slang and try to map it to the target language. Some \ac{SMT} systems like Google or Bing have recognition patterns over non-standard speech from old translations through the Web using \ac{SMT} approaches. In rare cases \ac{SMT} can solve this problem, but considering that new idiomatic expressions appear every day and most of them are isolated sentences, this challenge still remains open. Moreover, each person has their own speaking form.

Therefore, we suggest that user characteristics can be applied as context for solving the non-standard language problem. These characteristics can be extracted from social media or user logs and stored as user properties using \ac{SWT}, e.g., FOAF or SIOC~\cite{bojars2008using} vocabularies. These ontologies have properties which would help identify the birth place or the interests of a given user. For instance, the properties \emph{foaf:interest} and \emph{sioc:topic} can be used to describe a given person's topics of interest. If the person is a computer scientist and the model contains topics such as ``Information Technology" and ``Sports", the SPARQL queries would search for terms inserted in this context which are ambiguous. Furthermore, the property \emph{foaf:based\_near} may support the problem of idioms. Assuming that a user is located in a certain part of Russia and he is reading an English web page which contains some idioms, this property may be used to gather appropriate translations of idioms from English to Russian using a given RDF \ac{KB}\cite{moussallemlrec2018}. Therefore, an \ac{MT} system can be adapted to a user 
by using specific data about him in \ac{RDF} along with given \ac{KB}s. 


\subsection{Translating \ac{KB}s}


According to the surveyed articles, it is clear that \ac{SWT} may be used for translating \ac{KB}s in order to be applied in \ac{MT} systems. For instance, some content provided by the German Wikipedia version are not contained in the Portuguese one. Therefore, the semantic structure (i.e., triples) provided by DBpedia versions of these respective Wikipedia versions would be able to help translate from German to Portuguese. For example, the terms contained in triples would be translated to a given target language using a dictionary containing domain words. This dictionary may be acquired in two different ways. First, by performing localisation, as in the works by J. P. McCrae\cite{mccrae2016domain} and Arcan~\cite{arcan2013translating, arcan2013ontology} which translates the terms contained in a monolingual ontology, thus generating a bilingual ontology. Second, by creating embeddings of both DBpedia versions in order to determine the similarity between entities through their vectors. This insight is supported by some recent works, such as Ristoski et al.~\cite{ristoski2016rdf2vec}, which creates bilingual embeddings using \ac{RDF} based on Word2vec~\cite{mikolov2013efficient} algorithms, and Muhao Chen et al.~\cite{chen2017multi}, which generates multilingual knowledge graph embeddings for aligning entities across languages. Therefore, we suggest investigating an \ac{MT} approach mainly based on \ac{SWT} using \ac{NN} for translating \ac{KB}s. Once the \ac{KB}s are translated, we suggest including them in the language models for improving the translation of entities. 

Besides C. Shi et al~\cite{shi2016knowledge}, one of the recent attempts in this direction was carried out by Ar\v{c}an and Buitelaar~\citep{arcan2017translating}.\footnote{This paper was not included in this survey, because it was not peer-reviewed yet.} The authors aimed to translate domain-specific expressions represented by English \ac{KB}s in order to make the knowledge accessible for other languages. They claimed that \ac{KB}s are mostly in English, therefore they cannot contribute to the problem of \ac{MT} to other languages. Thus, they translated two \ac{KB}s belonging to medical and financial domains, along with the English Wikipedia, to German. Once translated, the \ac{KB}s were used as external resources in the translation of German-English. For the sake of brevity, they evaluated their approach in Phrase-based and \ac{NMT} systems. The results were quite appealing and the further research into this area should be undertaken.

\section{Conclusions and Future Work}
\label{sec:conc}
In this paper, we detailed a systematic literature review of \ac{MT} using \ac{SWT} for translating natural language sentences. The review aimed to answer the four research questions defined in~\autoref{sec:method} by a thorough analysis of the 21 most relevant papers.

Our goal was to provide a clear understanding of how \ac{SWT} have helped the translation process within \ac{MT} systems. Few studies have been found, suggesting that this method is still in its infancy. The surveyed articles demonstrate that \ac{SWT} have been mainly used for the disambiguation task in \ac{MT} systems and their capabilities have steadily increased. Considering the decision power of \ac{SWT}, they cannot be ignored by future \ac{MT} systems.

Nevertheless, there are still significant drawbacks. Although the research shows strong evidence of general \ac{SW} advantages in the translation process, as measured by automatic evaluation metrics, the real semantic contributions were assessed manually and the evaluators measured improvement according to the respective domains through which the work was approached. BLEU was the automatic evaluation metric used by all research works and it lacks semantic measurement, thus diminishing the real contribution of \ac{SW} in \ac{MT}. Recent works on \ac{MT} evaluation have shown promising advances for dealing with semantics in related metrics. These recent works~\cite{elloumi2015meteor,servan2016word2vec} attached \ac{SWT}, in this case DBnary~\cite{serasset2015dbnary}, to METEOR to handle meanings for non-English languages. In addition, MEANT has become the state-of-the-art in terms of evaluating semantics correlations in translated texts. Therefore, we expect that upcoming research works attempting to combine \ac{SWT} with \ac{MT} will be better evaluated by using these methods. 

\emph{How can \ac{SWT} enhance \ac{MT} quality?} Regarding this main research question, we identified that both semantic and syntactic disambiguation, including entities, structural di\-ver\-gen\-ce and \ac{OOV} words problems can all be tackled with \ac{SWT}. Additionally, \ac{SWT} and \ac{MT} approaches face the problem of languages being inherently ambiguous, in terms of lexicon, syntax, semantics and pragmatics. Identifying concepts in a \ac{KB} or finding the right translation for a word are instances of the same \ac{WSD} problem. Therefore, a deeper understanding of how pieces of information obtained from ontologies and \ac{KB}s on the one hand, and parallel and monolingual corpora on the other hand, may contribute solving these ambiguities in \ac{MT}. The two most recently surveyed articles,~\cite{shi2016knowledge,srivastava2017}, which have implemented a translation model relying on structured knowledge, support our conclusion. They showed promising results even using an automatic evaluation metric, which does not consider semantics. Therefore, this work increases the evidence that \ac{SWT} can significantly enhance the quality of \ac{MT} systems. 

Furthermore, we answered the first sub-question RQ1 by listing and discussing all surveyed articles. RQ2 is summarized in \autoref{directly} where we list all applied \ac{SW} methods and respective \ac{SW} resources. For RQ3, in \autoref{sec:mtsw} we also discussed the impact of applying an ontology to \ac{MT}. We conclude that ontological knowledge is generally beneficial for certain translation quality issues, mostly related to di\-sam\-bi\-gua\-tion in morphologically rich languages and sign languages. The ontological knowledge was a crucial resource for improving the translation between sign and spoken natural languages. We sought to answer RQ4, about the availability of {SW}-tools, and we could perceive that most of them were discontinued. Only AGDIS\-TIS~\cite{moussallem2017mag,Agdistis2014}, DBpedia Spot\-light\cite{spotlight} and Ba\-bel\-fy\cite{moro2014multilingual}, which had supported some of the surveyed articles, are currently available. Moreover, we noticed that these \ac{SW} tools performed only a specific task - disambiguation, for supporting the translations. Apart from the tools, BabelNet and DBpedia were used as \ac{KB}s by almost of all surveyed works.  

As a next step, we intend to elaborate a novel \ac{MT} approach based on our suggestions made in ~\autoref{sec:mtsw}. This involves an approach capable of simultaneously gathering knowledge from different \ac{SW} resources for addressing the ambiguity of named entities, which can also alleviate the problem of \ac{OOV} words. This insight relies on recent works \cite{chatterjee2017guiding,hokamp2017lexically}, which have guided the usage of external knowledge in \ac{NMT} systems for overcoming the vocabulary limitation of \ac{NN} models. Also, we aim to create a method to structure natural language sentences into triples for supporting the generation task. 

Future works that can be expected from fellow researchers, based on statements in their own papers, include the creation of linguistic ontologies describing the syntax of rich morphologically languages for supporting \ac{MT} approaches. In addition, alignment between ontologies is expected to try to bridge gaps that are not addressed by the current \ac{SMT} models. Since well-known bilingual dictionaries have been mapped to \ac{RDF}, the creation of multilingual dictionaries has become easier for content translation. These \ac{RDF} dictionaries can help to improve \ac{MT} steps, such as alignment, or even translate, based entirely on such semantic resources.

\section*{Acknowledgements}
This research has been partially supported by the Brazilian National Council for Scientific and Technological Development (CNPq) (no. 206971/2014-1) and the H2020 HOBBIT Project (GA No. 688227)


\bibliographystyle{elsarticle-num}
\bibliography{ref}

\end{document}